\documentclass[10pt,journal,compsoc]{IEEEtran}
\ifCLASSOPTIONcompsoc
  \usepackage[nocompress]{cite}
\else
  \usepackage{cite}
\fi
\ifCLASSINFOpdf
\else
\fi

\usepackage{amsthm}
\usepackage{amsfonts}
\usepackage{float}
\usepackage{graphicx}%
\usepackage{amsmath}
\usepackage{adjustbox}
\usepackage{subfigure}
\usepackage{wrapfig}  
\usepackage{algorithm}%
\usepackage{algorithmicx}%
\usepackage{algpseudocode}%

\usepackage{booktabs}
\usepackage{multirow}

\let\citet\cite

\begin{document}
\title{Accelerate Presolve in Large-Scale Linear Programming via Reinforcement Learning}
\author{
        Yufei~Kuang$^{\dagger}$,
        Xijun~Li$^{\dagger}$,
        Jie~Wang$^{\ast}$,~\IEEEmembership{Senior Member,~IEEE,}
        Fangzhou~Zhu,
        Meng~Lu,
        Zhihai~Wang,
        Jia~Zeng,~\IEEEmembership{Senior Member,~IEEE,}
        Houqiang~Li,~\IEEEmembership{Fellow,~IEEE,}
        Yongdong~Zhang,~\IEEEmembership{Senior Member,~IEEE,}
        and~Feng~Wu,~\IEEEmembership{Fellow,~IEEE}
\IEEEcompsocitemizethanks{
\IEEEcompsocthanksitem Y. Kuang, J. Wang, Z. Wang, H. Li, Y. Zhang, and F. Wu are with: a) CAS Key Laboratory of Technology in GIPAS, University of Science and Technology of China, Hefei 230027, China; b) Institute of Artificial Intelligence, Hefei Comprehensive National Science Center, Hefei 230091, China. 
E-mail: yfkuang@mail.ustc.edu.cn, jiewangx@ustc.edu.cn, zhwang@mail.ustc.edu.cn, lihq@ustc.edu.cn, zhyd73@ustc.edu.cn, fengwu@ustc.edu.cn. 
\IEEEcompsocthanksitem X. Li, F. Zhu, M. Lu, J. Zeng are with Huawei Noah’s Ark Lab. 
E-mail: xijun.li@huawei.com, zhufangzhou@huawei.com, lumeng22@huawei.com, Zeng.Jia@huawei.com.
}
\thanks{Manuscript received October, 2023. This work was done when the first author was an intern at Huawei Noah's Ark Lab. $^{\dagger}$Equal contribution. $^{\ast}$Corresponding author.}}

\markboth{Journal of \LaTeX\ Class Files,~Vol.~14, No.~8, August~2015}%
{Shell \MakeLowercase{\textit{et al.}}: Bare Demo of IEEEtran.cls for Computer Society Journals}

\IEEEtitleabstractindextext{
\begin{abstract}

Large-scale LP problems from industry usually contain much redundancy that severely hurts the efficiency and reliability of solving LPs, making presolve (i.e., the problem simplification module) one of the most critical components in modern LP solvers. 
However, how to design high-quality presolve routines---that is, the program determining (P1) which presolvers to select, (P2) in what order to execute, and (P3) when to stop---remains a highly challenging task due to the extensive requirements on expert knowledge and the large search space. 
Due to the sequential decision property of the task and the lack of expert demonstrations, we propose a simple and efficient reinforcement learning (RL) framework---namely, reinforcement learning for presolve (RL4Presolve)---to tackle (P1)-(P3) simultaneously. 
Specifically, we formulate the routine design task as a Markov decision process and propose an RL framework with adaptive action sequences to generate high-quality presolve routines efficiently. 
Note that adaptive action sequences help learn complex behaviors efficiently and adapt to various benchmarks. 
Experiments on two solvers (open-source and commercial) and eight benchmarks (real-world and synthetic) demonstrate that RL4Presolve significantly and consistently improves the efficiency of solving large-scale LPs, especially on benchmarks from industry. 
Furthermore, we optimize the hard-coded presolve routines in LP solvers by extracting rules from learned policies for simple and efficient deployment to Huawei's supply chain. 
The results show encouraging economic and academic potential for incorporating machine learning to modern solvers. 

\end{abstract}

}

\maketitle

\IEEEdisplaynontitleabstractindextext
\IEEEpeerreviewmaketitle

\IEEEraisesectionheading{\section{Introduction}\label{sec:introduction}}

\IEEEPARstart{L}{inear} programming (LP), which aims to optimize a linear objective subject to linear equality and inequality constraints, is one of the most fundamental model in mathematical optimization (MO) \cite{amp, ct, admm}. LP is widely used to formulate or approximate many important real-world optimization problems, e.g., network flow, routing, scheduling, and resource assignment \cite{ct, ce, gnetgen}, {in which the solving efficiency and solution quality are usually related to enormous economic value}. 
Moreover, LP serves as the cornerstone for other important MO models such as mixed-integer linear programming (MILP)\cite{reformulate}. 
Therefore, the optimization on LP solvers plays a central role in the development of many modern MO tools (e.g., Coin-OR, Gurobi, and OptVerse)  \cite{clp, gurobi, optverse}. 
Generally, we can write an LP problem in the following form \cite{ce}:
\begin{equation}\label{eq:definition}
    z^* = \min_{\mathbf{x}} \{ \mathbf{c}^{\top} \mathbf{x} \mid \underline{\mathbf{b}} \leq {A}\mathbf{x} \leq \overline{\mathbf{b}},  \underline{\mathbf{x}}\leq \mathbf{x}\leq\overline{\mathbf{x}} \}.
\end{equation}
Here $\mathbf{c}\in \mathbb{R}^n$ denotes the objective coefficient vector;  $\underline{\mathbf{x}}, \overline{\mathbf{x}}\in \mathbb{\bar{R}}^n, \underline{\mathbf{b}}, \overline{\mathbf{b}} \in \mathbb{\bar{R}}^{m}$ denote the variable and constraint bounds respectively, where $\mathbb{\bar{R}} = \mathbb{R}\cup  \{\pm\infty\}$; and ${A}\in \mathbb{R}^{m\times n}$ denotes the constraint coefficient matrix. The simplex and the interior-point algorithm are two mainstream algorithms to solve LP problem, and most modern solvers take them as the default LP algorithms \cite{admm}. 

\begin{figure*}[t]
\centering
\includegraphics[width=0.6\textwidth]{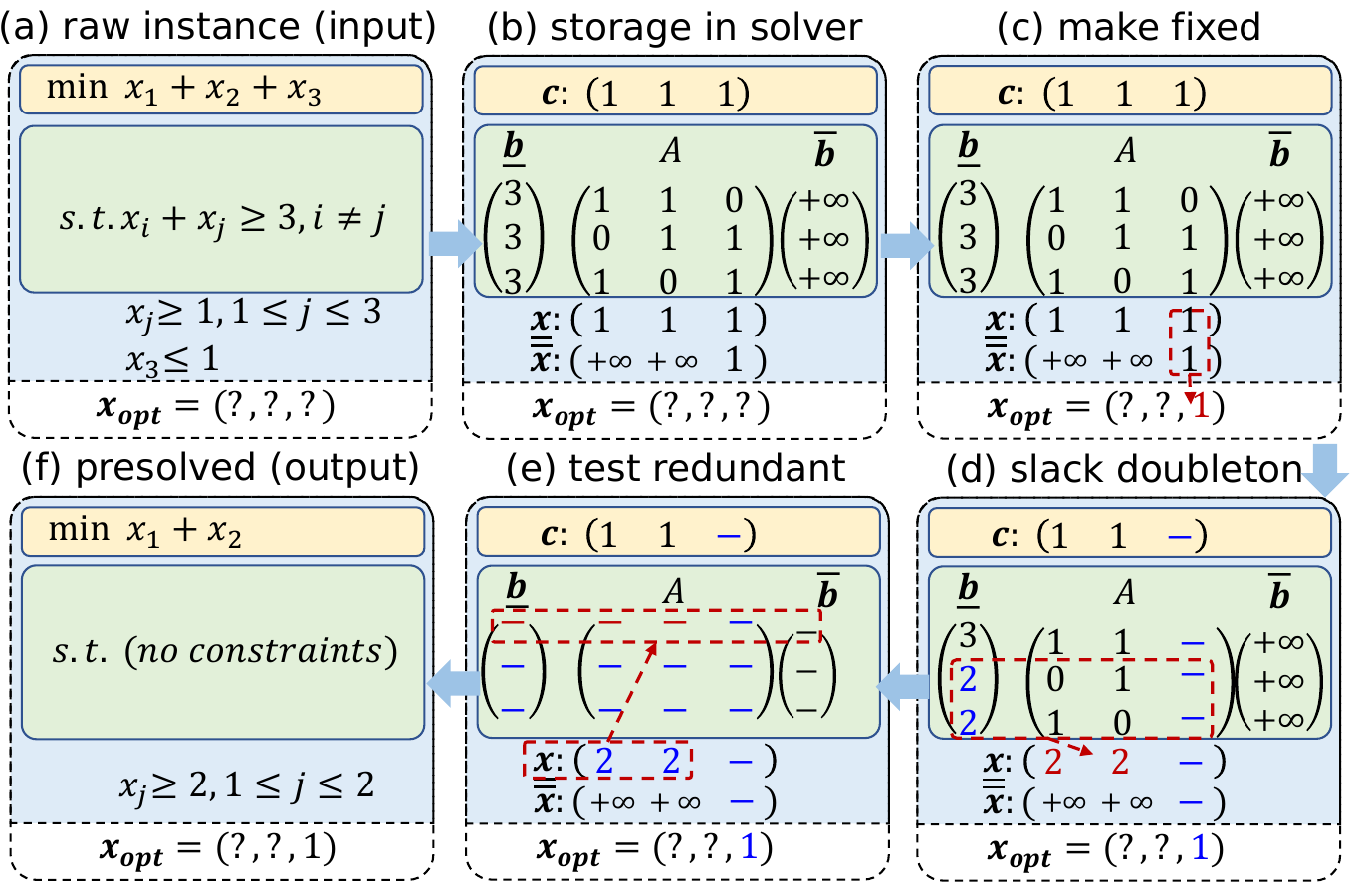}
\caption{Illustration the presolve process in modern LP solvers with a simple example. The solver first transforms the raw LP input to its general form in Equation (\ref{eq:definition}) and then executes different presolvers iteratively to remove the redundancy (see Table \ref{tab: presolvers} for descriptions for all presolvers). 
The red elements are modified by the current presolver and the blue ones are modified by previous presolvers. 
Note that the activation conditions of the current presolver (red box) usually rely on the presolve effect of previous ones (blue elements).}
\label{fig: example}
\end{figure*}

Presolve, which simplifies input LP problems by equivalent transformations before executing the LP algorithms mentioned above, is one of the most crucial components in modern LP solvers \cite{ct,ce}. 
LP problems, especially large-scale ones from industry, usually contain much redundancy. The redundancy comes from unprofessional modelings,  special structures, etc \cite{ct}, and it can severely decrease the efficiency and reliability of LP solvers to solve LPs \cite{ct} (see Table \ref{tab: routine-compare}). 
Thus, modern LP solvers integrate a rich set of presolvers to handle the redundancy from different aspects \cite{plp}. 
For example, the open-source LP solver Clp \cite{clp} (version 1.17) integrates fifteen different presolvers, among which the make fixed presolver fixes the value of variables whose lower bounds equal to their upper bounds. 
We illustrate the presolve process with a toy example in Figure \ref{fig: example} and list all presolvers of Clp with descriptions in Table \ref{tab: presolvers}.

Though it is widely recognized that the efficiency of LP solvers is integrally linked to the presolvers employed \cite{apt, ce}, we observe that the \textit{presolve routine}---that is, a sequence of presolvers successively executed in LP solvers---also {significantly} impacts the efficiency of solving large-scale LPs. 
In this paper, we conclude that an ideal presolve routine takes three core points into consideration. 
First, (P1) which presolvers to select? The effect of different presolvers varies greatly for different problems. Thus, selecting proper presolvers for specific problems usually improve the efficiency of presolve. 
Second, (P2) in what order to execute? Presolvers employed in modern solvers can affect each other \cite{ce}. Thus, a well-designed order for given presolvers can further improve the effect of presolve. 
Finally, (P3) when to stop? For example, the presolve process itself (and the corresponding postsolve process) can be time-consuming sometimes \cite{ct}. Thus, there is a trade-off between the presolve time and the pure LP solving time after presolve. 
In  Figure \ref{fig: analysis} and Table \ref{tab: routine-compare}, we empirically illustrate that (P1)-(P3) all play crucial roles in presolve routines.

However, how to design a presolve routine with better efficiency of solving LPs remains a highly challenging task, and previous research on this task is relatively limited. 
Currently, most modern LP solvers employ hard-coded presolve routines, in which all presolvers are executed in a static order with fixed number of iterations until no new reductions are found \cite{ce}. 
However, designing these routines usually requires much expert knowledge and perception of industrial data, which prevents researchers from academia to further involved. 
Moreover, hard-coded routines cannot capture distinct features for different inputs, and further optimization on them is challenging due to the large search space of all possible presolve routines.
In practice, problems collected from similar tasks usually share similar patterns \cite{ml4co, branch}. 
Based on this observation, recently, researchers apply machine learning (ML) approaches to components in MILP solvers like node selection \cite{ns}, variable selection \cite{branch}, and cut selection \cite{cut} to refine the algorithms. 
These studies achieve state-of-the-art performance on problems with chosen implicit distributions \cite{milp-survey, l2o-survey}, inspiring us to incorporate ML approaches to the presolve routine design task.

In this paper, we propose a novel approach---namely, reinforcement learning for presolve (RL4Presolve)---to design high-quality presolve routines automatically on given LP datasets. 
Specifically, there are three main components in RL4Presolve (see Figure  \ref{fig: illu} for illustration). 
First, we formulate the task as a reinforcement learning (RL) problem due to its sequential decision property and the lack of expert demonstrations. 
Then, to learn long presolve sequences with less cumulative time cost for decision making, we propose a novel adaptive action sequence that replaces primitive presolvers with automatically generated presolver sequences at each step. 
Intuitively, this approach is motivated from combos in video games \cite{openai-five} and sentence generation in natural language process \cite{s2s}. 
The appealing features of the adaptive action sequence include that it makes the agent efficient for complex behaviors, adaptive to various benchmarks, and more interpretable for deployment. 
Finally, we employ the proximal policy optimization algorithm \cite{ppo} to train the presolve agents efficiently. 
Experiments on two LP solvers and eight benchmarks demonstrate that RL4Presolve significantly and consistently improves the efficiency of solving LPs. 
Furthermore, we optimize the hard-coded presolve routines in LP solvers by extracting new rules from learned policies for simple and efficient deployment to Huawei’s supply chain, where one percent optimization on the objective can bring lots of dollars saving.

We summarize our major contributions as follows. 
(1) We observe from extensive experiments that the presolve routine plays a {critical} role in the efficiency of solving large-scale LPs and empirically conclude the three core points (P1)-(P3). (2) We propose a novel framework RL4Presolve with adaptive action sequences to tackle (P1)-(P3) {simultaneously} and {automatically}.
(3) We conduct extensive experiments to demonstrate that RL4Presolve consistently and significantly outperforms expert-designed presolve routines. (4) We propose a simple and efficient paradigm to deploy RL4Presolve to modern LP solvers and apply it to Huawei’s supply chain management system. (5) To the best of our knowledge, we are the first to formulate (P1)-(P3) in large-scale LP presolve, incorporate machine learning to tackle them, and deploy the learned rules to modern LP solvers for real-world applications.

\section{Related Work}\label{sec: related-work}
Machine learning for mathematical optimization (ML4MO) is a growing research field that replaces hand-crafted heuristics in MO algorithms to ML approaches \cite{ml4co}. 
Existing research can be roughly divided into two classes. One class assumes that empirical intuition exists and attempts to replace some heavy computations by fast approximations. 
For example, \citet{ns} uses given optimal solutions as oracle to learn node selection policies by imitation learning; \citet{kbranch} learns surrogate functions to replace the time-consuming strong branching strategy; and \citet{branch} and \citet{lb-branch} further employ graph neural networks to improve the performance. 
The other class assumes insufficient expert knowledge and thus use ML to automatically improve heuristics that are unsatisfactory yet. 
For example, \citet{heuristic} learn to schedule the time-consuming heuristics in MILP solvers  to reduce the primal integral, \citet{cut} and \citet{wangcut} learn to select cuts to improve the lower bound of MILP problems; \citet{cg} learn to select columns in the column generation algorithm; and \citet{reformulate} reformulate the input LP problems to improves the solving efficiency. All the four works mentioned above employ RL to train the policies. 
Generally, it is a natural idea to apply RL in this class of research due to the absence of expert demonstrations. 
The goal of this paper lies in the second class, i.e., to improve the hard-coded presolve routines. 
Thus, we employ RL as the training algorithm. 

Routine optimization is a common and critical issue that widely appears in the development of modern mathematical optimization solvers. Thus, there are two recent researches, proposed by \citet{heuristic} and  \citet{wangcut}, whose goals are relatively similar to that in this paper. 
Specifically, \citet{heuristic} propose a novel data-driven heuristic  to schedule heuristics in mixed integer linear programming (MILP) solvers. Note that presolve routine design task in our paper is an infinite horizon sequential decision task because all presolvers can be executed repeatedly. Thus, directly applying the formulation in \citet{heuristic} to design a heuristic for presolve routine is intractable. \citet{wangcut} propose a hierarchical sequence model for cut sequence selection in MILP solvers. 
Intuitively, this task is similar to the convex hull task handled in \cite{pn}, as they both select an ordered subset of featured elements from the original set. However, the presolve routine design task in this paper is somewhat similar to the sentence generation task, as each presolver is a primitive action like a separate word.

\section{Preliminaries} \label{sec: preliminaries}
We briefly introduce the preliminaries in this section.

\begin{figure}[t]
\centering
\includegraphics[width=0.48\textwidth]{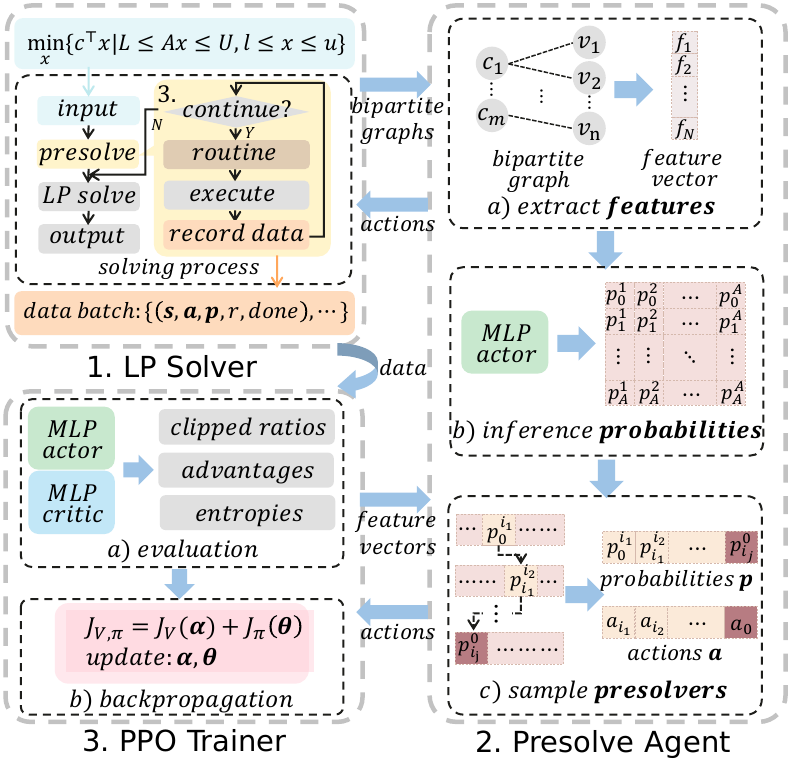}
\caption{Illustrate all modules in RL4Presolve. Here $N,A$ are the length of features and the total number of presolvers (see Appendix \ref{sec: preliminaries} for details), $p_i^j = \pi_{\boldsymbol{\theta}}(i\mid j, \mathbf{s})$ is the conditional probability from $i$th presolver to $j$th, and $a_0=E$ is the end token.}
\label{fig: illu}
\end{figure}

\textbf{Presolve in LP Solvers}
LP problems obtained from real-world applications usually contains much redundant information \cite{plp}, which comes from unprofessional modeling for the optimization problems, special structure for specific problems, etc. 
The redundancy can take different forms in practice, e.g., multiple constraints (rows) that are linear dependent with each other or a single variable (column) whose value can be fixed in advance \cite{apt, amp}. 
Then, the presolve component in an LP solver integrates various presolvers to simplify the input LP problems from different aspect accordingly, e.g., removing redundant constraints, fixing the value of variables, and tightening bounds for constraints or variables \cite{ce}. 
These presolvers reduce the number of nonzero elements (NNZ) of the input problems by detecting and removing the redundancy and ultimately improve the solving efficiency \cite{ct}. 
For example, \citet{pt} show a reduction of variables by $20\%$ and the total solving time by $10\%$ on the LP datasets they test. We further test on three real-world LP datasets from  the industry to illustrate the significant effect of presolve in Table \ref{tab: routine-compare}. 
We illustrate the presolve process via a vanilla example in Figure \ref{fig: illu} and list all presolvers in Clp (an open-source LP solver) with descriptions in Table \ref{tab: presolvers}. 

\textbf{Markov Decision Process (MDP)} An MDP is defined by a tuple $(\mathcal{S}, \mathcal{A}, p, r, \gamma)$ \cite{rlbook}. Here $\mathcal{S}$ is the state space; $\mathcal{A}$ is the action space;  $p:\mathcal{S} \times \mathcal{A} \rightarrow \Delta_{\mathcal{S}}$ is the transition probability function, where $\Delta_{\mathcal{S}}$ is the set of probability measures on $\mathcal{S}$; 
$r:\mathcal{S}\times\mathcal{A}\rightarrow \mathbb{R}$ is the reward function; and $\gamma \in [0,1]$ is the discount factor. 
Define $\pi: \mathcal{S} \rightarrow \Delta_\mathcal{A}$ as the policy, then a reinforcement learning (RL) agent optimizes the policy $\pi$ by maximizing the cumulative rewards \cite{regularized_rl}.

\begin{figure*}[t]
    \centering
    \subfigure[NNZ reduction for presolvers.]{
        \label{fig: p1}
        \includegraphics[width=0.30\textwidth]{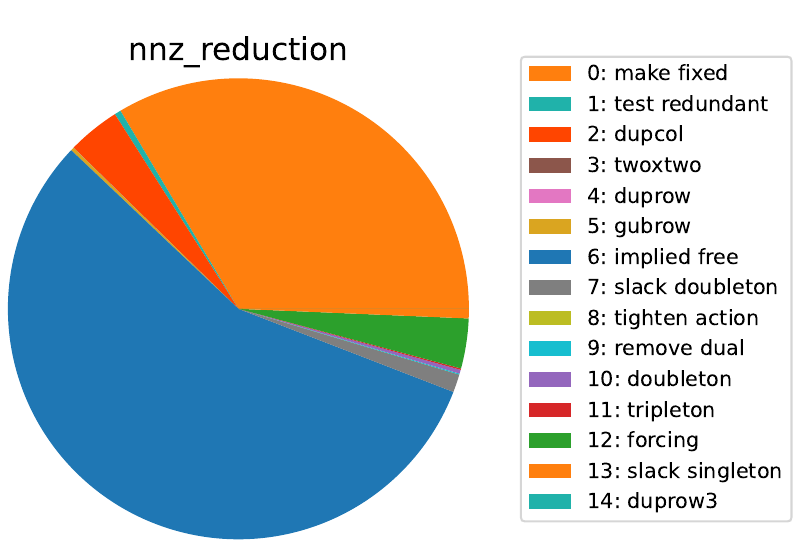}}
    \subfigure[Effects and activation conditions.]{
        \label{fig: p2}
        \includegraphics[width=0.33\textwidth]{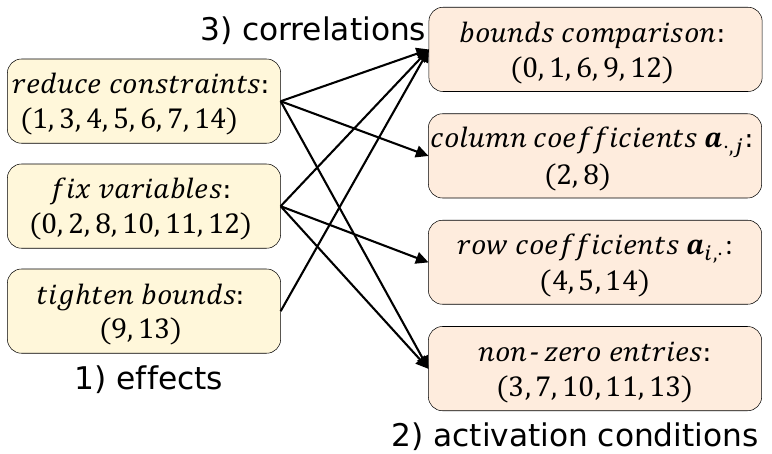}}
     \subfigure[Presolve process visualization.]{
        \label{fig: p3}
        \includegraphics[width=0.31\textwidth]{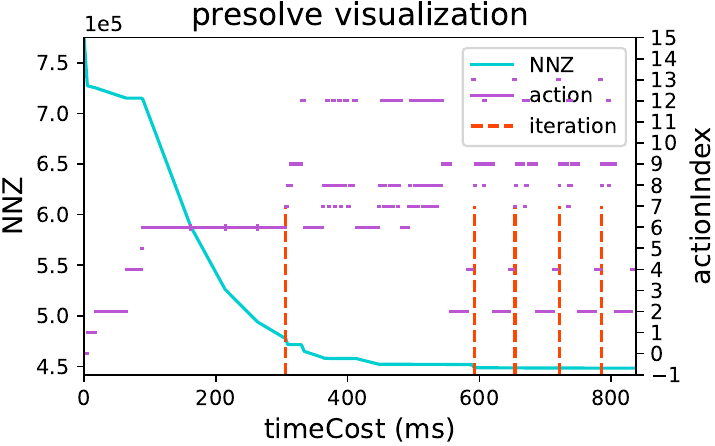}}
\caption{Illustrate (P1)-(P3) on the Production Planning benchmark on Clp. Figure \ref{fig: p1} shows that the effect of each presolver varies greatly. Figure \ref{fig: p2} sorts all presolvers by their effects and activation conditions and then visualizes their one-step correlations. Figure \ref{fig: p3} records the iterations, the selected presolvers, and the NNZ changes to show the gradually decreased effect during presolve. }
\label{fig: analysis}
\end{figure*}

\section{Problem Setup} \label{sec: setup}
Though it is widely recognized that presolve significantly impacts the efficiency of solving LP problems, most previous researches mainly focus on designing and analyzing specific presolvers \cite{amp}. However, during the development of our Optverse solver, we observe that: a) presolve routines play \textit{crucial} roles in the efficiency of solving LPs, and b) designing well-performed presolve routines is a \textit{non-trivial} task. 
In this section, we introduce the presolve routine design task and analyze it in detail for problem setup. 
To make the results more instructive and reproducible for real-world applications, all experiments here are conducted with three real-world benchmarks from different scenarios and tasks in the advanced planning and scheduling system (APS) \cite{aps} of Huawei's supply chain on Clp. 

\textbf{What is a Presolve Routine?} 
A presolve routine determines a sequence of presolvers successively executed in the presolving phase of integrated LP solving process. 
Currently, most modern solvers use hard-coded presolve routines, in which all presolvers are executed in a static order with fixed number of iterations until no new reductions are found, for all instances \cite{ce}. 
These iterative, in-order routines were first presented by \citet{amp} and remain in use over the 30 years later \cite{ce}. 
See Algorithm 12 in \citet{ce} for descriptions of the presolve routine of Clp as an example.

\textbf{What matters in LP Presolve Routines?} We conduct experiments from different aspects to understand that. Based on the results, we conclude that a well-designed presolve routine is supposed to take the three points into consideration simultaneously:
\begin{enumerate}
    \item[(P1)] \textit{Which presolvers to select?} Though a number of presolvers are integrated in an LP solver, not all of them are both effective and efficient. 
    As shown in Figure \ref{fig: p1} and Figure \ref{fig: p1-all},  the time cost and the effect (i.e., the NNZ reduction for input problems) for different presolvers vary greatly. Thus, simply executing all presolvers (as most LP solvers do) for all instances only results in limited performance due to the inefficient presolvers. 
    
    \item[(P2)] \textit{In what order to execute?} Presolvers employed in LP solvers can affect each other. For example, presolvers that remove columns also change the NNZ of rows, which may then activate other presolvers (see Figure \ref{fig: example}). We visualize some correlations of presolvers in Clp using a directed graph in Figure \ref{fig: p2}. Another result to support this claim explicitly can be found in Table 3.1 in \citet{ce}. Most modern solvers employ manually designed orders whose design usually require much empirical intuition and expert knowledge.
    
    \item[(P3)] \textit{When to stop?} Though more iterations usually result in better simplification on input instances, the presolve process itself (and the corresponding postsolve process) can be time-consuming. Thus, there is a trade-off between the presolve time and the pure LP solving time. 
    Figure \ref{fig: p3} and Figure \ref{fig: p3-all} visualize the NNZ curve of a randomly selected instance, where the presolve process continues for a long time with the effect gradually decreasing. 
\end{enumerate}
We further design three different types of presolve routines and compare them with the default one in Clp. Specifically, the top/last $k\%$ routines only use the top/last $k\%$ presolvers in terms of NNZ reductions (manually selected based on statistical results of each instance), the reordering routines sort the presolvers of the default routine randomly in each iteration, and the iteration $\pm k \%$ routines increase/decrease the iteration number in default routine. Results in Table \ref{tab: routine-compare} show that all the three points above in a presolve routine play a critical role in the efficiency of solving  LPs.

\textbf{What Makes the Routine Design Task Non-Trivial?} 
In practice, we find that designing high-quality presolve routines is a non-trivial task due to two main challenges. 
\textit{First}, the space of possible presolver sequences grows exponentially with respect to the length of sequence. Thus, either optimizing existing routines or designing new ones are intractable with traditional approaches. 
\textit{Second}, we observe that the optimal routines for different LP instances are different (see Figure \ref{fig: analysis-all} for comparisons between different benchmarks). Thus, simply applying a static presolve routine for all inputs cannot capture the distinct features for different inputs. 
Moreover, manually designed presolve routines---which are widely employed in modern LP solvers---usually require much empirical intuition about different presolvers and their correlations. 
This knowledge heavily relies on  the perception of industrial data and thus prevents  researchers from academia to further involving\footnote{Currently, presolve routines in most commercial LP solvers are closed source.}. 

\begin{table*}[t]
\caption{Comparison of time, presolve time, and NNZ reduction on three different benchmarks, seven presolve routines, and 256 instances for each benchmark on Clp. Results show that (P1)-(P3) all impact the efficiency of solving LPs in real-world applications.}
\centering
\begin{adjustbox}{width=0.99\textwidth}
\small
\begin{tabular}{cccccccccc}
\toprule\toprule
cDataset:             & \multicolumn{3}{c}{Master Production Schedule}& \multicolumn{3}{c}{Production Planning} & \multicolumn{3}{c}{Supply and Demand Matching
}  \\ \cmidrule(r){1-4} \cmidrule(lr){5-7} \cmidrule(l){8-10}
Method       & Time ($s$)               & Presolve time ($s$)   & NNZ reduction ($\%$)  & Time ($s$)              & Presolve time ($s$)   & NNZ reduction ($\%$)  & Time ($s$)               & Presolve time ($s$)   & NNZ reduction ($\%$)  \\ \cmidrule(r){1-4} \cmidrule(lr){5-7} \cmidrule(l){8-10}
default      & 2.98                    & 2.55                 & 95.92                & 4.97                   & 2.42                 & 48.41                & 19.03                   & 5.19                 & 48.63                \\
presolve-off & 3.28 (+10.1\%)          & NA                   & NA                   & 45.21 (+809.1\%)       & NA                   & NA                   & 104.25 (+447.8\%)       & NA                   & NA                   \\ \cmidrule(r){1-4} \cmidrule(lr){5-7} \cmidrule(l){8-10}
top $60\%$      & 2.65 (-11.1\%)          & 2.24 (-12.1\%)       & 95.90 (-0.0\%)       & 10.58 (+112.7\%)       & 1.36 (-43.9\%)       & 45.49 (-6.0\%)       & 16.82 (-11.6\%)         & 4.33 (-16.6\%)       & 50.15 (+3.1\%)       \\
last $60\%$       & 3.10 (+3.9\%)           & 0.94 (-62.9\%)       & 20.55 (-78.6\%)      & 27.42 (+451.3\%)       & 1.03 (-57.5\%)       & 12.52 (-74.1\%)      & 107.07 (+462.6\%)       & 0.92 (-82.3\%)       & 0.17 (-99.6\%)       \\ \cmidrule(r){1-4} \cmidrule(lr){5-7} \cmidrule(l){8-10}
reordering  & 3.23 (+8.4\%)           & 2.76 (+8.3\%)        & 95.91 (-0.0\%)       & 5.55 (+11.6\%)         & 3.11 (+28.3\%)       & 48.49 (+0.2\%)       & 19.15 (+0.6\%)          & 5.03 (-3.2\%)        & 50.13 (+3.1\%)       \\ \cmidrule(r){1-4} \cmidrule(lr){5-7} \cmidrule(l){8-10}
iteration $-40\%$  & 2.52 (-15.6\%)          & 2.09 (-17.9\%)       & 95.78 (-0.1\%)       & 4.66 (-6.2\%)          & 2.12 (-12.6\%)       & 48.41 (-0.0\%)       & 19.56 (+2.8\%)          & 4.60 (-11.5\%)       & 48.04 (-1.2\%)       \\
iteration $+40\%$  & 3.18 (+6.5\%)           & 2.76 (+8.4\%)        & 96.06 (+0.1\%)       & 5.69 (+14.4\%)         & 3.14 (+29.6\%)       & 48.44 (+0.1\%)       & 18.95 (-0.4\%)          & 7.31 (+40.7\%)       & 52.56 (+8.1\%)       \\ \bottomrule
\end{tabular}
\end{adjustbox}
\label{tab: routine-compare}
\end{table*}

\textbf{What is the Goal of This Paper?} 
Machine learning (ML) approaches excel at handling high-dimensional tasks and automatically extracting features \cite{ml4co}. In this paper, we incorporate ML approaches to presolve components so as to a) design high-quality presolve routines automatically and b) optimize the hard-coded routines employed in LP solvers based on the learned ones.

\section{Methods} \label{sec: method}
The presolve routine design is a sequential decision task and few expert demonstrations are available. Thus, it is natural to formulate it as an RL \cite{rlbook} task. 
In this section, we  propose a simple and efficient RL framework with adaptive action sequences to generate high-quality presolve sequences {efficiently}.

\subsection{Reinforcement Learning Formulation} \label{sec: rl-formulation}
We start with specifying the components of the MDP and the objective. See the execution process mentioned in this section in Figure \ref{fig: illu} Part 1. 
More discussions on the choice of state space and rewards can be referred in Appendix \ref{sec: discussion-mdp}. 

\textbf{The State Space} 
Though a bipartite graph $\boldsymbol{g}$ contains most information about an LP problem \cite{gnn-lp}, using a smaller subset of information that is relevant to the downstream task can make the training easier.
Thus, we define a lightweight  {feature vector} $\boldsymbol{s}$ (more precisely, an observation vector) containing  compressed information for presolve. 
We use that since a) designing graph neural networks \cite{graph} and tuning the training hyperparameters for large-scale graphs to extract features is usually challenging \cite{gnn-design}; b) the activation condition of each presolver is clearly described in literature, so we can design required features easily.
See Table \ref{tab: features} for descriptions of features.

\textbf{The Action Space} 
We define the action space as the set of all the {presolver sequences}. Specifically, define $\mathcal{A}_0$ as the set of all available presolvers, then $\mathcal{A} = \left\{ \boldsymbol{a} \mid \boldsymbol{a} = (a_i)_{i=1}^n,\,a_i\in \mathcal{A}_0, n\in \mathbb{N} \right\}$. When $n=0$ the action $\boldsymbol{a}$ reduces to an empty presolver sequence. We define it as the action to terminate presolve. Obviously, $\mathcal{A}_0$ is isomorphic to a small subset of $\mathcal{A}$. However, the larger $\mathcal{A}$ significantly increases the possible actions at each step,  enabling the agent to learn more sophisticated behaviors. We will explain the motivation why such definition is necessary in Section \ref{sec: agent}.

\textbf{The Environment Transition} 
We regard the LP solver as a black box where the transition function is unknown. 
At each step, the agent interacts with the solver to obtain a feature vector and determines a presolver sequence to execute next. 
If the agent decides to continue presolving, then the solver executes the selected presolvers and returns a presolved LP instance to compute a new feature vector. Otherwise, the solver terminates presolve and returns final results after the solving process finished.

\textbf{The Reward Function and the Discount Factor}
We set the reward function $r=-t$ and the discount factor $\gamma=1$, where $t$ is the executing time between two steps. Thus, the absolute value of the cumulative rewards from initial states to terminal states equals to the total solving time of the LPs.

\textbf{The Objective} We run the solver repeatedly with a policy $\pi(\boldsymbol{a}\mid \boldsymbol{s})$ and LP instances sampled from distribution $p_0(\boldsymbol{s})$. 
The probability of a trajectory $\tau = (\boldsymbol{s}_0, \boldsymbol{a}_0, \boldsymbol{s}_1, \boldsymbol{a}_1, \cdots, \boldsymbol{s}_{T-1}, \boldsymbol{a}_{T-1}, \boldsymbol{s}_T)$ is
\begin{equation}
p_\pi(\tau)=p\left(\mathbf{s}_0\right) \prod_{t=0}^{T-1} \pi\left(\boldsymbol{a}_t \mid \mathbf{s}_t\right) p\left(\mathbf{s}_{t+1} \mid \mathbf{s}_t, \boldsymbol{a}_t\right),
\end{equation}
where $\boldsymbol{s}_T$ is the terminal state after receiving an empty presolver sequence $\boldsymbol{a}_{T-1}$. Then, the RL agent learns the routine design policy $\pi$ by maximizing the  following cumulative rewards:
\begin{equation} \label{eq: obj}
J(\pi)=\mathbb{E}_{\pi}\left[\sum_{t=0}^{T-1} r\left(\boldsymbol{s}_t, \boldsymbol{a}_t\right)\right].
\end{equation}

\begin{table*}[t]
\caption{Suppose the vanilla RL model converges to policies similar to RL4Presolve and compare their cumulative decision time. The results show that the adaptive action sequence significantly reduces the cumulative time cost for decision making.}
\centering
\begin{adjustbox}{width=0.9\textwidth}
\small
\begin{tabular}{lcccc}
\toprule\toprule
Benchmarks                   & \begin{tabular}[c]{@{}c@{}}      Time cost\\per decision($s$)    \end{tabular} & \begin{tabular}[c]{@{}c@{}}  Presolver \\ Number     \end{tabular} & \begin{tabular}[c]{@{}c@{}}    Total decision time \\ vanilla RL($s$)     \end{tabular}    & \begin{tabular}[c]{@{}c@{}}    Total decision time\\ RL4Presolve($s$)      \end{tabular} \\ \midrule
Master   Production Schedule & 0.041                       & 165.14                                 & 6.78                                   & 0.26                                                      \\
Production Planning          & 0.025                       &{49.60}              & 1.24                                   & 0.025                                                     \\
Supply Demand Matching       & 0.036                       & 153.37                                 & 5.52                                   & 0.25                                                      \\
Generalized Network Flow     & 0.027                       & 90.44                                  & 2.44                                   & 0.32                                                     \\ \bottomrule
\end{tabular}
\end{adjustbox}
\label{tab: decision-time}
\end{table*}

\subsection{Agent with Adaptive Action Sequence} \label{sec: agent}
Now we consider how to design an RL-based presolve agent. We illustrate the agent described in this section in Figure \ref{fig: illu} Part 2.

\textbf{Trade-Off between Decision Time and Quality} 
Compared to traditional RL algorithms, the objective in Equation (\ref{eq: obj}) suggests us to explicitly consider the decision time. 
If we only select a single presolver for each decision, then the cumulative decision time can be extremely expensive due to relatively high time costs for each decision. 
However, if we formulate the task as a contextual bandit \cite{rlbook} and make a decision only once to obtain the whole presolve routine, then learning long sequences is challenging, which can result in limited performance.

\textbf{Adaptive Action Sequence} To tackle this challenge, we propose a novel approach named adaptive action sequence, which replaces primitive presolvers with automatically generated presolver sequences at each step. 
Instead of selecting single presolvers from $\mathcal{A}_0$, we define probabilities for all presolver sequences in $\mathcal{A}$ and then sample from it. 
Recall that the success rate of the latter presolvers is directly affected by the former. 
Thus, we leave the long-term dependency implicitly to different states, and use the chain rule to simplify the probability model by only considering one-step dependency, i.e., 
\begin{align}\label{eq: action-probability}
\begin{aligned}
p(\boldsymbol{a} \mid \mathbf{s}) &=  p\left(a_1 \mid \mathbf{s}\right)\prod_{i=2}^{n+1} p\left(a_i \mid a_1,\cdots,a_{i-1}, \mathbf{s}\right) \\
&= p\left(a_1 \mid \mathbf{s}\right)\prod_{i=2}^{n+1} p\left(a_i \mid a_{i-1}, \mathbf{s}\right) .
\end{aligned}
\end{align}
Here $\boldsymbol{a}=(a_i)_{i=1}^n$ is a sequence containing $n$ presolvers, $a_{n+1}=E$ is a special token to represent the end of sequences,  $p\left(a_1 \mid \mathbf{s}\right)$ is the initial distribution, and $p\left(a_i \mid a_{i-1}, \mathbf{s}\right)$ is the one step transition probability. 
We parameterize these probabilities by function approximators like neural networks, i.e.,
\begin{equation} \label{eq: policy}
    \pi_{\boldsymbol{\theta}}(\boldsymbol{a} \mid \mathbf{s}) = \pi_{\boldsymbol{\theta}}\left(a_1 \mid \mathbf{s}\right)\prod_{i=2}^{n+1} \pi_{\boldsymbol{\theta}}\left(a_i \mid a_{i-1}, \mathbf{s}\right) .
\end{equation}

\textbf{Understanding the Proposed Method} The proposed method employs two steps. The first step is to reduce the decision frequency via sequential actions, which is motivated from combos in video games \cite{openai-five}. 
However, designing combos for presolve is difficult due to the lack of human knowledge. 
Thus, the second step is to generate ``combos'' automatically. This step is similar to sentence generation tasks \cite{s2s}, while there are two distinctions. 
First, to reduce the decision time, we only inference once and then sample a whole sequence from the fixed transition probabilities. 
Second, we only model the one-step dependency into the transition, as the long-term dependency is implicitly included in the states. More precisely, we model the presolver sequence generation at each step as a  parameterized Markov chain \cite{prob}.

\textbf{The Appealing Features} There are three appealing features for the adaptive action sequence. \textit{First}, compared to the vanilla RL model, it reduces the cumulative time cost for decision making. Consider the RL4Presolve learned policies in Table \ref{tab: routine-rule}, suppose the vanilla RL model converges to policies similar to that, then the cumulative decision time can be relatively long (see Table \ref{tab: decision-time}). 
\textit{Second}, compared to the bandit model, the adaptive action sequence is more flexible, making it easier to adapt to various datasets. 
On the MIRPLIB benchmark, the learned policy executes more than $200$ presolvers on average for each instance (see Table \ref{tab: routine-rule}), which is challenging for the bandit model to learn at once. On the Supply Demand Matching benchmark, poor presolve severely decreases the solving efficiency (see Table \ref{tab: routine-compare}), while the bandit model forces to execute presolve for only one turn, which results in very poor presolve effect of the randomly initialed bandit agent and eventually extremely long training time. On the contrary, adaptive action sequences work well and consistently on all benchmarks. 
\textit{Finally}, the adaptive action sequence explicitly models the one-step dependency for all presolvers, making the learned policies easier to visualize and understand.

\begin{algorithm}[t]
	\caption{Reinforcement Learning for Presolve (RL4Presolve)}
	\label{alg: rl4presolve} 
	\begin{algorithmic}
	\State \textbf{Input:} 
        actor $\pi_{\boldsymbol{\theta}}$; critic $V_{\boldsymbol{\alpha}}$; learning rate $\beta$; total iterations $M$; total collected samples $S$ and training epoches $N$ per iteration.
		\For{iteration$=1,2,\cdots, M$}
            \State \textit{// Run the LP solver to collect samples:}
            \State Initial data buffer: ${D}\leftarrow\emptyset$.
            \While{size$({D}) < S$}
		\State Sample presolver sequences: $\boldsymbol{a}\sim \pi_{\boldsymbol{\theta}}(\cdot \mid \mathbf{s})$.
            \State Collect samples: ${D}\leftarrow {D} \cup \left\{\left( 
\mathbf{s}, \boldsymbol{a}, \pi_{\boldsymbol{\theta}_{\text{old}}}, -\Delta t, \text{done} \right)\right\}$.
            \State Update states: $\mathbf{s}^\prime\sim p(\cdot\mid \mathbf{s}, \boldsymbol{a}), \mathbf{s}\leftarrow\mathbf{s}^\prime$.
            
            \EndWhile
            \State \textit{// Train the presolve agent via PPO algorithm:}
		\For{epoch$=1,2,\cdots,N$}
            \State Sample minibatch $\mathcal{B}\sim {D}$ to calculate $J_V(\boldsymbol{\alpha}), J_\pi(\boldsymbol{\theta})$.
            \State Update: $\boldsymbol{\alpha}\leftarrow\text{Adam}(\boldsymbol{\alpha}, \nabla_{\boldsymbol{\alpha}}J_V), \boldsymbol{\theta}\leftarrow\text{Adam}(\boldsymbol{\theta}, \nabla_{\boldsymbol{\theta}}J_\pi)$.
		\EndFor
		\EndFor
	\State \textbf{Output:}  actor $\pi_{\boldsymbol{\theta}}$
	\end{algorithmic}
\end{algorithm}

\subsection{Training Algorithm with Action Sequence} \label{sec: ppo}

We employ proximal policy optimization (PPO) \cite{ppo} to train the parameterized transition probabilities above.
See the pseudo code in Algorithm \ref{alg: rl4presolve} and the illustration in Figure \ref{fig: illu} Part 3. 

Denote $ r(\boldsymbol{\theta})=\frac{\pi_{\boldsymbol{\theta}}\left(\boldsymbol{a} \mid \boldsymbol{s}\right)}{\pi_{\boldsymbol{\theta}_{\text {old }}}\left(\boldsymbol{a} \mid \boldsymbol{s}\right)} $ as the probability ratio of the current and the old policies. 
Then, PPO clips $r(\boldsymbol{\theta})$ in its objective to avoid excessively large policy updates \cite{ppo, ppo-study}, i.e., 
\begin{equation}\label{eq: ratio}
r_\epsilon^{\text{clip}}(\boldsymbol{\theta})=\begin{cases}
1+\epsilon, &\hat{A}>0 \text{ and } r(\boldsymbol{\theta})\geq1+\epsilon; \\
1-\epsilon, &\hat{A}<0 \text{ and } r(\boldsymbol{\theta})\leq1-\epsilon;\\
r(\boldsymbol{\theta}), &\text{otherwise}.
\end{cases}
\end{equation}
Here $\epsilon$ is a hyperparameter for clip and $\hat{A}$ is an estimator of the advantage \cite{rlbook}. Then, PPO maximizes:
\begin{equation}\label{eq: ppo}
J_\pi(\boldsymbol{\theta})  = \hat{\mathbb{E}}_t\left[ r_\epsilon^{\text{clip}}(\boldsymbol{\theta})\hat{A} \right],
\end{equation}
where $\hat{\mathbb{E}}_t[\cdot]$ is the empirical average over current samples. 

There are two reasons why PPO is preferred. 
First, designing a state-action function (i.e., the $Q$ function) compatible with the countable infinite set $\mathcal{A}$ is challenging. Policy based algorithms avoid that by training the parameterized policies directly using a performance objective \cite{rlbook}. 
Second, due to the large search space and the time-consuming solving process (see Table \ref{tab: routine-results}), The training algorithm is required to be both sample efficient and easy to parallelize.

Based on Equation (\ref{eq: policy}), the probability ratio can be written as:
\begin{equation}\label{eq: unclipped-ratio}
r(\boldsymbol{\theta})= \frac{\pi_{\boldsymbol{\theta}}\left(a_1 \mid \mathbf{s}\right)\prod_{i=2}^{n+1} \pi_{\boldsymbol{\theta}}\left(a_i \mid a_{i-1}, \mathbf{s}\right) }{\pi_{\boldsymbol{\theta}_{\text{old}}}\left(a_1 \mid \mathbf{s}\right)\prod_{i=2}^{n+1} \pi_{\boldsymbol{\theta}_{\text{old}}}\left(a_i \mid a_{i-1}, \mathbf{s}\right) } = \prod_{i=1}^{n+1} \frac{\pi_{i}(\boldsymbol{\theta})}{\pi_{i}(\boldsymbol{\theta}_{\text{old}})},
\end{equation}
where $\pi_{1}(\boldsymbol{\theta})=\pi_{\boldsymbol{\theta}}(a_1\mid \boldsymbol{s})$ and $\pi_{i}(\boldsymbol{\theta})=\pi_{\boldsymbol{\theta}}(a_i\mid a_{i-1}, \boldsymbol{s})$ for $ i\geq 2$. 
We also use a state-value function $V_{\boldsymbol{\alpha}}(\boldsymbol{s})$ as a baseline to reduce the variance of the advantage-function estimator \cite{ppo}. For simplicity of implementation, we use the finite-horizon estimator, i.e., 
\begin{equation}\label{eq: advantage}
\hat{A} = R(\boldsymbol{s}) - V_\alpha(\boldsymbol{s}).
\end{equation}
Then, we train the state-value function $V_{\boldsymbol{\alpha}}(\boldsymbol{s})$ by minimizing
\begin{equation}
J_V(\boldsymbol{\alpha}) = \hat{\mathbb{E}}_t\left[ \frac{1}{2} \hat{A}^2 \right]. 
\end{equation}
More details about the training settings and the hyperparameters can be found in Appendix \ref{sec: more-implementations}.

\section{Experiments} \label{sec: exp}

We conduct extensive experiments to evaluate RL4Presolve, which mainly have five goals: 
a) to illustrate that RL4Presolve improves the efficiency of solving LPs significantly and consistently; 
b) to test its generalization ability to different LP algorithms and larger instances;
c) to analyse the learned policies and employ to modern LP solvers.
d) to show that RL4Presolve tackles (P1)-(P3) simultaneously; 
e) to conduct ablation study on the adaptive sequence proposed in Section \ref{sec: method}. Unless mentioned, all experiments here are conduct on the open-source LP solver Clp \cite{clp} by default.

\textbf{Benchmarks} We evaluate RL4Presolve on eight benchmarks, consisting of four real-world and four synthetic. 
Benchmark 1-3: Master Production Schedule, Production Planning, and Supply and Demand Matching are three real-world LP benchmarks from different scenarios and tasks in the advanced planning and scheduling system (APS) \cite{aps} of Huawei’s supply chain \cite{supply-chain1, supply-chain2}. 
Benchmark 4: MIRPLIB \cite{mirp} is the LP relaxation of an open-source real-world MILP dataset on maritime inventory routing problems. 
Benchmark 5-7: Facility Location \cite{facilities}, Set Covering \cite{setcover}, and Multicommodity Network Flow \cite{fcm} are three open-source synthetic MILP benchmarks widely used in previous research. We generate instances larger than that in \citet{branch, node-compare} in our experiment as solving the LP relaxations is usually much faster. 
Benchmark 8: Generalized Network Flow \cite{gnetgen} is an open-source synthetic LP benchmark. 
We report the size of different benchmarks and the hyperparameters we used to generate them in Table \ref{tab: benchmark-hyperparameters} in Appendix. 

\textbf{The Insight for the Benchmark Selection} The eight benchmarks used in Section \ref{sec: exp} are selected based on the following two insights. \textit{First}, we select benchmarks that are relatively sparse. As claimed by the most well-known study from \citet{plp}, a presolve process might not be advantageous if the constraint matrix $A$ is not sparse. Thus, in the  synthetic benchmarks, we manually set their hyperparameters to make the instances sparse enough. \textit{Second}, we select benchmarks that are relatively complex for solving. Note that even with the proposed adaptive action sequences, the RL agent still requires more than 20ms per decision for feature extraction and network inference (see Table \ref{tab: decision-time}). Thus, if the instances are too simple to solve, then the acceleration obtained by the RL agent can be slight due to the time cost for decision. The optimization on these simple instances is also unnecessary. 
We note that these two insights are usually satisfied in practice, as many LP problems from real-world applications are sparse and large-scale.

\textbf{Baselines} We apply RL4Presolve to Clp \cite{clp} (the open-source LP solver developed by COIN-OR Foundation) and OptVerse \cite{optverse} (the commercial solver developed by Huawei). 
Due to the lack of previous research, we implement two enhanced baselines that significantly outperform the default routines on benchmarks from Huawei's supply chain, which are motivated from both our expert knowledge and the analysis on Table \ref{tab: routine-compare}. 
Specifically, Enhance-v1 employs the best $40\%$ presolvers in terms of the reductions of number of non-zero elements (NNZ) and disables the other $60\%$ on each benchmark. Intuitively, this is also a data-driven baseline since best presolvers for different benchmarks are very different. Enhance-v2 reduces the default number of iterations by $40\%$.

\textbf{Training and Evaluation Settings} For the MIRPLIB benchmark, we use $70\%$, $5\%$, and $ 25\%$ of the total datasets for training, validation, and test due to the limited number of total available instances. For all the other benchmarks, we use $1000$, $16$, and $64$ instances, respectively. During training, we sample instances repeatedly and then take the sampled instance as an environment to train the RL agent. We employ the dual simplex algorithm---which is usually the default LP algorithm in many modern solvers \cite{clp, gurobi}---as the LP algorithm after presolve. After every ten training iterations, we record the performance of current policy on the validation set. We select the best-performed policy on the validation set and evaluate it on the test set. 
We train four policies for each benchmark with different random seeds to make the results more convincing. 
We tune all hyperparameters on the OptVerse LP solver and the Production Planning benchmark and then directly apply them to Clp and all the other benchmarks. 
More details for implementation can be found in Appendix \ref{sec: more-implementations}.

\begin{table*}[t]
\caption{Compare RL4Presolve to three baselines on two LP solvers and eight benchmarks. We report mean value of time with standard deviation on four seeds. Results show that RL4Presolve \textit{significantly} and  \textit{consistently} improves the efficiency of solving LPs. }
\centering
\begin{adjustbox}{width=1.0\textwidth}
\small
\begin{tabular}{cccccccccccccc}
\toprule
\toprule
\multicolumn{2}{c}{Dataset:} & \multicolumn{3}{c}{Master Production   Schedule}          & \multicolumn{3}{c}{Production Planning}                      & \multicolumn{3}{c}{Supply Demand Matching}                & \multicolumn{3}{c}{MIRPLIB}                                \\ 
\cmidrule(r){1-5} \cmidrule(lr){6-8} \cmidrule(lr){9-11} \cmidrule(l){12-14}
Solver      & Method         & Time($s$)              & Improvement(\%) & Wins(\%)       & Time($s$)                 & Improvement(\%) & Wins(\%)       & Time($s$)              & Improvement(\%) & Wins(\%)       & Time($s$)               & Improvement(\%) & Wins(\%)       \\
\cmidrule(r){1-5} \cmidrule(lr){6-8} \cmidrule(lr){9-11} \cmidrule(l){12-14}
Clp         & Default        & 2.65(±5.1\%)           & NA              & 0           & 6.15(±0.3\%)              & NA              & 0           & 20.80(±0.5\%)          & NA              & 4.69           & 126.37(±0.1\%)          & NA              & 27.27          \\
            & Enhance-v1     & 2.04(±1.5\%)           & 23.21           & 6.25           & 4.76(±0.6\%)              & 22.60           & 22.66           & 14.52(±0.3\%)          & 30.19           & 41.41          & 125.08(±0.1\%)          & 1.02            & 18.18          \\
            & Enhance-v2     & 2.31(±2.4\%)           & 12.98           & 0           & 5.79(±0.1\%)              & 5.83            & 0           & 16.25(±0.1\%)          & 21.88           & 6.64           & 128.57(±0.1\%)          & -1.74           & 2.73           \\
            & RL4Presolve    & \textbf{1.58(±6.5\%)}  & \textbf{40.63}  & \textbf{93.75} & \textbf{4.26(±3.6\%)}     & \textbf{30.72}  & \textbf{77.34} & \textbf{14.17(±8.8\%)} & \textbf{31.88}  & \textbf{47.27} & \textbf{107.35(±1.0\%)} & \textbf{15.05}  & \textbf{51.82} \\
\cmidrule(r){1-5} \cmidrule(lr){6-8} \cmidrule(lr){9-11} \cmidrule(l){12-14}
OptVerse    & Default        & 3.70(±3.0\%)           & NA              & 0           & 5.99(±0.5\%)              & NA              & 0           & 17.89(±1.7\%)          & NA              & 16.41          & 115.79(±0.1\%)          & NA              & 8.33           \\
            & Enhance-v1     & 3.10(±0.4\%)           & 16.20           & 10.16          & 3.83(±0.3\%)              & 35.98           & 9.38          & 13.32(±0.6\%)          & 25.54           & 9.77           & 106.22(±0.5\%)          & 8.26            & 25.00          \\
            & Enhance-v2     & 3.81(±1.5\%)           & -2.97           & 0           & 5.64(±0.8\%)              & 5.80            & 0           & 17.43(±1.7\%)          & 2.57            & 11.72          & 104.14(±0.4\%)          & 10.06           & 5.00           \\
            & RL4Presolve    & \textbf{2.02(±7.9\%)}  & \textbf{45.41}  & \textbf{89.84} & \textbf{2.81(±1.3\%)}     & \textbf{53.04}  & \textbf{90.63} & \textbf{10.12(±5.0\%)} & \textbf{43.44}  & \textbf{62.11} & \textbf{96.98(±2.6\%)}  & \textbf{16.25}  & \textbf{61.67} \\ 
\bottomrule
\end{tabular}
\end{adjustbox}
\newline
\vspace{1mm}
\newline
\begin{adjustbox}{width=1.0\textwidth}
\small
\begin{tabular}{cccccccccccccc}
\toprule
\toprule
\multicolumn{2}{c}{Dataset:} & \multicolumn{3}{c}{Facility Location}                     & \multicolumn{3}{c}{Set Covering}                             & \multicolumn{3}{c}{Multicommodity Network Flow}           & \multicolumn{3}{c}{Generalized Network Flow}               \\
\cmidrule(r){1-5} \cmidrule(lr){6-8} \cmidrule(lr){9-11} \cmidrule(l){12-14}
Solver      & Method         & Time($s$)              & Improvement(\%) & Wins(\%)       & Time($s$)                 & Improvement(\%) & Wins(\%)       & Time($s$)              & Improvement(\%) & Wins(\%)       & Time($s$)               & Improvement(\%) & Wins(\%)       \\
\cmidrule(r){1-5} \cmidrule(lr){6-8} \cmidrule(lr){9-11} \cmidrule(l){12-14}
Clp         & Default        & 27.98(±0.7\%)          & NA              & 3.13           & 16.64($\pm0.4\%$)         & NA              & 0              & 4.62(±2.4\%)           & NA              & 9.38           & 1.15(±1.1\%)            & NA              & 0.39           \\
            & Enhance-v1     & 26.95(±3.0\%)          & 3.62            & 22.66          & 16.55($\pm0.7\%$)         & 0.5             & 0              & 4.49(±3.5\%)           & 2.83            & 20.70          & 1.12(±1.5\%)            & 2.61            & 23.83          \\
            & Enhance-v2     & 27.25(±0.5\%)          & 2.60            & 17.19          & 16.41($\pm1.3\%$)         & 1.34            & 0              & 4.77(±4.4\%)           & -3.27           & 10.16          & 1.10(±1.4\%)            & 4.35            & 31.64          \\
            & RL4Presolve    & \textbf{25.64(±1.1\%)} & \textbf{8.33}   & \textbf{57.03} & \textbf{1.94($\pm1.8\%$)} & \textbf{88.36}  & \textbf{100}   & \textbf{4.08(±6.3\%)}  & \textbf{11.74}  & \textbf{59.77} & \textbf{1.08(±7.6\%)}   & \textbf{6.09}   & \textbf{44.14} \\
\cmidrule(r){1-5} \cmidrule(lr){6-8} \cmidrule(lr){9-11} \cmidrule(l){12-14}
OptVerse    & Default        & 14.67(±2.3\%)          & NA              & 11.33          & 1.18($\pm0.3\%$)          & NA              & 0              & 2.24(±0.5\%)           & NA              & 0              & 1.21(±0.3\%)            & NA              & 14.06          \\
            & Enhance-v1     & 14.58(±1.6\%)          & 0.63            & 16.80          & 1.12($\pm1.4\%$)          & 4.92            & 21.48          & 2.05(±0.8\%)           & 8.28            & 0              & 1.13(±0,3\%)            & 6.60            & 28.52          \\
            & Enhance-v2     & 15.09(±1.0\%)          & -2.83           & 13.67          & 1.15($\pm0.5\%$)          & 1.99            & 2.34           & 2.17(±1.5\%)           & 3.08            & 0              & 1.25((±0,5\%)           & -3.31           & 8.59           \\
            & RL4Presolve    & \textbf{12.97(±3.4\%)} & \textbf{11.60}  & \textbf{58.20} & \textbf{1.08($\pm1.2\%$)} & \textbf{7.93}   & \textbf{76.17} & \textbf{1.36(±4.3\%)}  & \textbf{39.45}  & \textbf{100}   & \textbf{1.04(±5.7\%)}   & \textbf{14.05}  & \textbf{48.83} \\ \bottomrule
\end{tabular}
\end{adjustbox}
\label{tab: routine-results}
\end{table*}

\begin{table*}[t]
\caption{Test generalization ability of trained policies to different LP algorithms and larger instances. Results show that RL4Presolve generalizes to both primal simplex algorithm and larger instances.}
\centering
\begin{adjustbox}{width=0.9\textwidth}
\small
\begin{tabular}{cccccccccc}
\toprule
\toprule
\multicolumn{2}{c}{Dataset:} & \multicolumn{2}{c}{Production Planning}         & \multicolumn{2}{c}{Supply Demand Matching}          & \multicolumn{2}{c}{Multicommodity Network   Flow}       & \multicolumn{2}{c}{Facility Location}         \\
\cmidrule(r){1-4} \cmidrule(lr){5-6} \cmidrule(lr){7-8} \cmidrule(l){9-10}
LP Algorithm   & Method      & Time($s$)                  & Improvement(\%)    & Time($s$)                   & Improvement(\%)       & Time($s$)                      & Improvement(\%)        & Time($s$)                 & Improvement(\%)   \\
\cmidrule(r){1-4} \cmidrule(lr){5-6} \cmidrule(lr){7-8} \cmidrule(l){9-10}
Primal Simplex & Default     & 7.72(±1.2\%)               & NA                 & 180.43(±3.1\%)                        & NA                    & 1.22(±7.8\%)                   & NA                     & 102.19(±16.0\%)           & NA                \\
               & RL4Presolve & \textbf{5.49(±5.3\%)}      & \textbf{28.25}     & \textbf{150.88(±6.3\%)}             & $\mathbf{16.18}$ & \textbf{0.58(±3.6\%)}          & \textbf{52.34}         & \textbf{93.11(±20.6\%)}   & \textbf{8.89}     \\
\cmidrule(r){1-4} \cmidrule(lr){5-6} \cmidrule(lr){7-8} \cmidrule(l){9-10}
Interior Point & Default     & \textbf{130.47(±1.0\%)}    & NA                 & \textbf{171.37  (±5.1\%) }                     & NA                    & 3.07(±1.1\%)                   & NA                     & 269.99(±2.3\%)            & NA                \\
               & RL4Presolve & 137.39(±6.2\%)             & -5.3               & {176.71(±7.4\%)}             & {${-3.12}$}  & \textbf{2.37(±3.1\%)}          & \textbf{22.69}         & \textbf{264.94(±2.5\%)}   & \textbf{1.87}     \\
\bottomrule
\end{tabular}
\end{adjustbox}
\newline
\vspace{1mm}
\newline
\begin{adjustbox}{width=0.9\textwidth}
\small
\begin{tabular}{cccccccccc}
\toprule
\toprule
\multicolumn{2}{c}{Dataset:} & \multicolumn{2}{c}{Production Planning (Large)} & \multicolumn{2}{c}{Supply Demand Matching (Large)}  & \multicolumn{2}{c}{Multicommodity Network Flow (Large)} & \multicolumn{2}{c}{Facility Location (Large)} \\
\cmidrule(r){1-4} \cmidrule(lr){5-6} \cmidrule(lr){7-8} \cmidrule(l){9-10}
Solver         & Method      & Time($s$)                  & Improvement(\%)    & Time($s$)                   & Improvement(\%)       & Time($s$)                      & Improvement(\%)        & Time($s$)                 & Improvement(\%)   \\
\cmidrule(r){1-4} \cmidrule(lr){5-6} \cmidrule(lr){7-8} \cmidrule(l){9-10}
Clp            & Default     & 51.17(±1.1\%)              & NA                 & 83.27($\pm3.2\%$)           & NA                    & 17.56(±4.9\%)                  & NA                     & 130.34(±5.4\%)            & NA                \\
               & RL4Presolve & \textbf{40.27(±5.9\%)}     & \textbf{21.30}     & \textbf{67.59($\pm5.8\%$)}  & \textbf{18.83}        & \textbf{15.92(±0.6\%)}         & \textbf{9.35}          & \textbf{117.05(±9.3\%)}   & \textbf{10.2}     \\
\cmidrule(r){1-4} \cmidrule(lr){5-6} \cmidrule(lr){7-8} \cmidrule(l){9-10}
OptVerse       & Default     & 36.43(±3.1\%)              & NA                 & 61.00($\pm4.7\%$)           & NA                    & 5.71(±0.1\%)                   & NA                     & 79.03(±7.5\%)             & NA                \\
               & RL4Presolve & \textbf{25.96(±6.7\%)}     & \textbf{28.74}     & \textbf{50.63($\pm10.8\%$)} & \textbf{16.99}        & \textbf{3.28(±1.0\%)}          & \textbf{42.54}         & \textbf{68.77(±11.3\%)}   & \textbf{12.99}   \\
\bottomrule
\end{tabular}
\end{adjustbox}
\label{tab: routine-transfer}
\end{table*}

\begin{table*}[t]
\caption{Details of the solving process. Here time is the total solving time, LP time is the pure LP solving time after presolve, and presolver number is the total executed presolvers. The total time is slightly larger than presolve time add LP time due to other modules like postsolve in modern  LP solvers. The results help us to further understand what RL4Presolve learned on each benchmark.}
\centering
\begin{adjustbox}{width=0.95\textwidth}
\small
\begin{tabular}{ccccccccccc}
\toprule
\toprule
Dataset:        & \multicolumn{4}{c}{Production Planning}                          & \multicolumn{1}{l}{} & \multicolumn{5}{c}{Multicommodity Network   Flow}                                      \\
\cmidrule(r){1-6}  \cmidrule(l){7-11}
Method          & Time($s$) & Presolve Time($s$) & LP Time($s$) & Presolver Number & NNZ Reduction($\%$)  & Time($s$) & Presolve Time($s$) & LP Time($s$) & Presolver Number & NNZ Reduction($\%$) \\
\cmidrule(r){1-6}  \cmidrule(l){7-11}
Default         & 6.15      & 2.74               & 2.66         & 260.62           & \textbf{49.65}                 & 4.62      & 1.01               & 3.12         & 35.00            & 33.33                \\
RL4Presolve     &\textbf{ 4.26}      & 0.98               & 2.79         & 49.60            & 48.32                 & 4.08      & 0.44               & 3.16         & 1.28             & 33.33                \\
Extracted Rules & 4.40      & 0.42               & 3.66         & 24               & 46.62                 & \textbf{3.71}      & 0.17               & 3.11         & 2                & 33.33                \\
\bottomrule
\end{tabular}
\end{adjustbox}
\newline
\vspace{1mm}
\newline
\begin{adjustbox}{width=0.95\textwidth}
\small
\begin{tabular}{ccccccccccc}
\toprule
\toprule
Dataset:        & \multicolumn{5}{c}{MIRPLIB}                                                             & \multicolumn{5}{c}{Set Covering}                                                       \\
\cmidrule(r){1-6}  \cmidrule(l){7-11}
Method          & Time($s$) & Presolve Time($s$) & LP Time($s$) & Presolver Number & NNZ Reduction($\%$)  & Time($s$) & Presolve Time($s$) & LP Time($s$) & Presolver Number & NNZ Reduction($\%$) \\
\cmidrule(r){1-6}  \cmidrule(l){7-11}
Default         & 126.37    & 0.08               & 125.79       & 95.82            & 7.38                 & 16.64     & 0.09               & 16.54        & 34.78            & \textbf{1.68 }               \\
RL4Presolve     & \textbf{107.35}    & 0.51               & 106.65       & 254.45           & \textbf{7.62}                 & 1.94      & 0.03               & 1.91         & 0.21             & 0.01                \\
Extracted Rules & 114.95    & 0.22               & 114.60       & 157.36           & 7.47                 & \textbf{1.90}      & NA                 & 1.88         & NA               & NA                 \\
\bottomrule
\end{tabular}
\end{adjustbox}
\label{tab: routine-rule}
\end{table*}

\begin{figure*}[t]
    \centering
    \subfigure[Improved presolve efficiency.]{
        \label{fig: analyse-policy: p1}
        \includegraphics[width=0.35\textwidth]{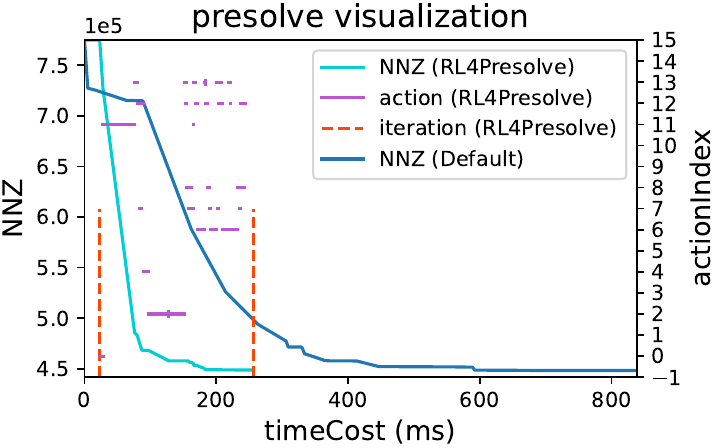}}
    \subfigure[Effect of learned orders.]{
        \label{fig: analyse-policy: p2}
        \includegraphics[width=0.28\textwidth]{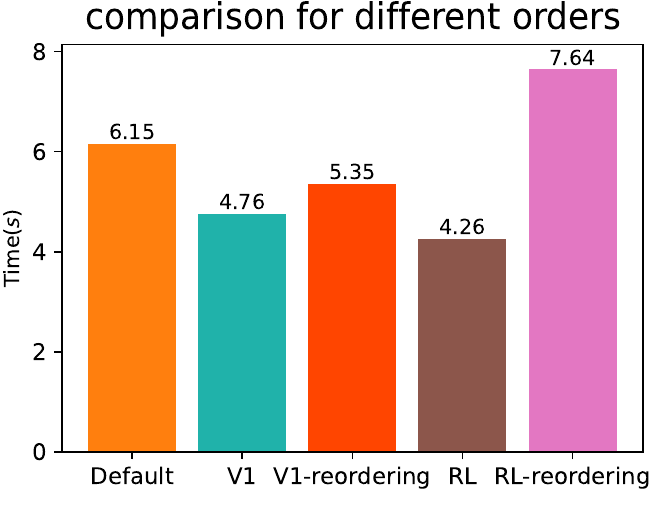}}
    \subfigure[Presolve and LP time trade-off.]{
        \label{fig: analyse-policy: p3}
        \includegraphics[width=0.31\textwidth]{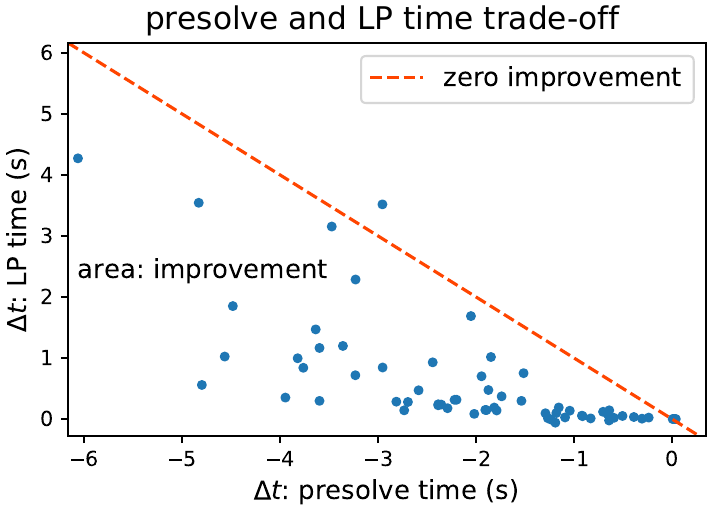}}
\caption{Illustrate that RL4Presolve tackles (P1)-(P3) simultaneously on Production Planning. }
\label{fig: analyse-policy}
\end{figure*}

\subsection{Comparative Evaluation}  \label{sec: exp-evaluation}

\textbf{Solving Efficiency} We compare RL4Presolve to different baselines on eight benchmarks and report the results in Table \ref{tab: routine-results}. Results show that RL4Presolve \textit{significantly} and \textit{consistently} improves the efficiency of solving LPs. 
Note that RL4Presolve achieves such improvement by \textit{only} changing the default presolve routines. 
We observe that the improvement on real-world benchmarks is usually significant, which matches the prior experience that instances from real-world applications usually contain more redundancy due to their unprofessional modelings. 
The standard deviations on many benchmarks are small, which is because we use $64$ instances for evaluation over each seed, and the asymptotic performance of the stochastic policies may have specific upper bounds (though the corresponding optimal policies may not be unique). 
We further report the total environment steps and the corresponding training time for each benchmark in Table \ref{tab: env-step}.

\textbf{Generalization Ability} We test the generalization ability of RL4Presolve trained policies to different downstream LP algorithms and larger instances. 
First, we apply the learned presolve policies to the primal simplex and the interior-point algorithms \cite{ct, cvx}, which are also widely deployed in modern LP solvers.  Then, we test learned policies on larger benchmarks (see Table \ref{tab: benchmark-hyperparameters} for descriptions). Results in Table \ref{tab: routine-transfer} shows that policies trained with dual simplex algorithm \textit{generalize well} to both the primal simplex algorithm and larger instances. However, the generalization ability of trained policies to the interior-point algorithm is poor. A potential reason to explain that is the large difference between these two LP algorithms, e.g., the difference between the pure LP solving time of these two LP algorithms makes their best time to stop presolve (P3) totally different.

\subsection{Applications to Modern Solvers} \label{sec: exp-application}
Due to the complex deployment of neural networks and the hardware (GPU) constraints, applying ML techniques to modern solvers directly is usually challenging. 
However, the visualization of adaptive action sequences helps us further understand what RL4Presolve learned. 
In this section, we illustrate how we optimize the hard-coded presolve routines in Clp based on analysis on the learned policies. 

\textbf{Analysis on What RL4Presolve Learned} We visualize all the learned policies and observe that RL4Presolve tends to generate \textit{similar} action probabilities for instances from similar distributions (see Figure \ref{fig: policy-similarity} for an example). To further understand what RL4Presolve learned, we report four case studies in Table \ref{tab: routine-rule} on a)  Production Planning, b) Multicommodity Network Flow, c) MIRPLIB, and d) Set Covering. 
For a) and b), we observe that the agent tends to output presolver sequences for only one turn and then terminates presolve immediately. Moreover, for a), RL4Presolve slightly decreases the NNZ reduction but obtains more acceleration on presolve; for b), RL4Presolve achieves comparable presolve effect with much fewer presolvers. 
For c), interestingly, the agent repeatedly outputs almost random presolver sequences (i.e., the agent outputs nearly uniform distributions as transition probabilities) at all steps until it terminates presolve with a small probability, while this simple policy outperforms the default one (However, simply increasing the presolve iterations in the default routines does not help, as many solvers like Clp force to terminate presolve if little new reductions are found in several steps.). We conclude that this is because more presolve can effectively reduce the solving time while presolve itself is fast on this benchmark. 
For d), on the contrary, the agent turns off presolve directly even if presolve can reduce the NNZ of input instances, as solving presolved instances is more time-consuming than solving the original instances directly on this benchmark. A potential reason is that presolve simplifies the problems but destroys some crucial structure in this benchmark, which hurts the performance of solving presolved problems afterwards.

\textbf{Optimize the Presolve Routine of Clp} Analysis above motivates us that on each benchmark we may extract some new routines from learned policies that outperforms the default one. 
Therefore, we sample $20$ presolve routines on each benchmark from learned policies and then replace the default routine with the best-performed one. 
We find this approach \textit{consistently} improves the hard-coded presolve routines. 
We visualize the learned policies in Figure \ref{fig: policy-visualize} and report the improved performance of extracted rules in Table \ref{tab: routine-rule}. Note that extracted rules do not require \textit{any} additional hardwares like GPUs, which means the proposed paradigm is simple and efficient for deployment in practice. 

\textbf{Discussions} Observe that the extracted rules above are relatively simple to understand, so what is the necessity to incorporating RL to presolve routine design? \textit{First}, RL4Presolve defines a large enough search space for complex presolve routines and provides an efficient training approach to find high-quality routines for different benchmarks {adaptively}. \textit{Second}, the similarity discovered by RL among instances from similar tasks gives important clues that designing the presolve routine benchmark-wisely (rather than instance-wisely) can also be effective sometimes. Thus, future optimizations on presolve routines may employ simpler techniques (e.g., black-box parameter tuning tools) directly on each benchmark for {faster} trials. If the performance is unsatisfactory, then RL4Presolve is applied. 
\textit{Finally}, the learned policies give us insights on how to manually optimize presolve for different benchmarks, which can be quite valuable when data or hardware for data-driven methods is limited.

\subsection{Analysis and Ablation Study} \label{sec: ablation}

\textbf{Analysis on (P1)-(P3)} 
We conduct experiments on Production Planning to show that RL4Presolve tackles (P1)-(P3) simultaneously and report the results in Figure \ref{fig: analyse-policy}. 
For (P1), Figure \ref{fig: analyse-policy: p1} visualizes the presolve process of a randomly selected problem, in which RL4Presolve selects presolvers that reduce the NNZ more efficiently than the default rule. 
For (P2), Figure \ref{fig: analyse-policy: p2} compares the performance of Enhance-v1 and RL4Presolve to their corresponding randomly reordered versions, which demonstrates that order matters in these routines. 
For (P3), Figure \ref{fig: analyse-policy: p3} visualize the presolve time change versus LP (after presolve) time change between RL4Presolve and the default rule to illustrate the trade-off, i.e., the agent turns off presolve earlier to reduce presolve time on this benchmark. 
Another result to support these claims is in Table \ref{tab: routine-rule}: For (P1), the learned policies reduce the NNZ with much fewer presolvers executed on the first two benchmarks. For (P3), the learned policies increase presolve time or close presolve directly on the last two benchmarks. 

\begin{figure*}[t]
    \centering
    \subfigure[Ablation study on the adaptive action sequence.]{
        \label{fig: ablation}
c        \includegraphics[width=0.35\textwidth]{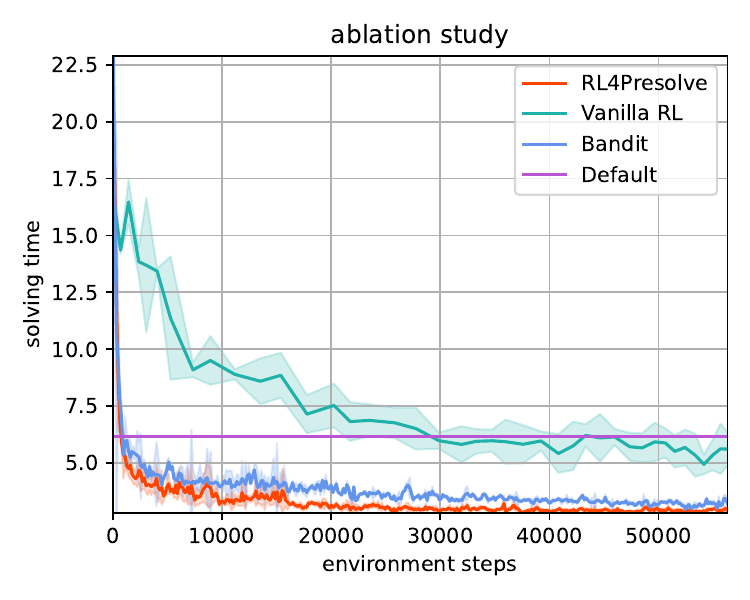}}
    \subfigure[Sensitive analysis on learning rates.]{
        \label{fig: sensitivity}
        \includegraphics[width=0.35\textwidth]{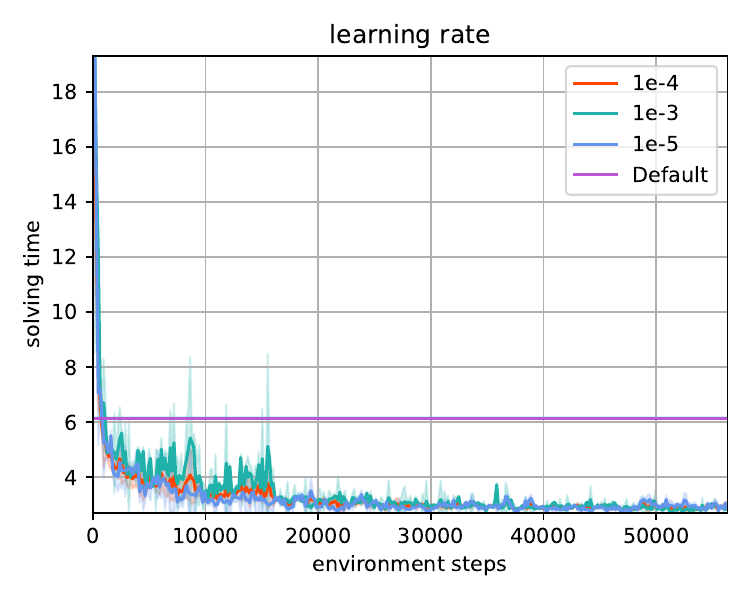}}
\caption{Ablation study and sensitivity analysis for RL4Presolve. Results show that the adaptive action sequence plays a critical role for training (see the ablation study part for detailed discussions) and the training process is insensitive to different learning rates.}
\label{fig: curves}
\end{figure*}

\textbf{Ablation Studies for Adaptive Action Sequences} 
We employ two additional agents that select a single presolver for each decision (vanilla RL, 1) and select all presolvers at once (bandit, 2). 
Results in Figure \ref{fig: ablation} show that: 
(1) The vanilla RL spends much time on decision making, making its total solving time longer than RL4Presolve. 
(2) Though the bandit model achieves performance similar to RL4Presolve on Production Planning, it takes nearly \textit{five} times longer for training. This is because the bandit model forces the agent to only execute presolve for one turn, making the initial presolve effect to be extremely poor, and finally long solving time. 
The total training time on Supply Demand Matching is even too long to obtain and report. 
Moreover, bandit model is not compatible to benchmarks like MIRPLIB where the total required presolvers is large. 
Thus, we note that the adaptive action sequence can adapt to various benchmarks that are very different in characteristics. See Appendix \ref{app: exp} for more results.

\textbf{Sensitivity Analysis on Learning Rate} We evaluate RL4Presolve with different learning rate $\beta$ to analyze its sensitivity and report the training curves in Figure \ref{fig: sensitivity}. 
Results show that the training stability of RL4Presolve is relatively insensitive to $\beta$, which explains why the hyperparameters tuned on OptVerse and the Production Planning benchmark can directly apply to all the solvers and benchmarks with consistent performance.

\section{Conclusion}
In this paper, we propose the first ML framework that optimizes the hard-coded presolve routines in modern solvers to improve the efficiency of solving large-scale linear programming (LP) problems, and we apply the rules extracted from learned policies to modern LP solvers for simple and efficient deployment. Experimental results demonstrate that RL4Presolve significantly and consistently improves the efficiency of solving large-scale LPs. We believe that our work shows {encouraging} economic and academic potential for incorporating machine learning to modern solvers.

\ifCLASSOPTIONcompsoc
  \section*{Acknowledgments}
\else
  \section*{Acknowledgment}
\fi

The authors would like to thank all the anonymous reviewers for their insightful comments. This work was supported by National Key R\&D Program of China under contract  2022ZD0119801, National Nature Science Foundations of China grants U19B2026, U19B2044, 61836011, 62021001, and 61836006, and the Fundamental Research Funds for the Central Universities grant WK3490000004.

\bibliographystyle{IEEEtran}
\bibliography{refs}

\clearpage
\newpage

\appendices

\renewcommand\thefigure{\Alph{section}\arabic{figure}}   
\setcounter{figure}{5}    
\renewcommand\thetable{\Alph{section}\arabic{table}}    
\setcounter{table}{3}

\section{Discussions}

\textbf{Motivations} There are two preliminary observations motivating us to study on LP presolve routine design. 
First, large-scale LPs from industry usually contain much redundancy that severely decreases the solving efficiency. 
Second, the presolve process for large-scale LPs is usually both complex and time-consuming, which indicates the potential of optimization on this component. 
We incorporate ML to presolve as ML approaches excel at handling complex tasks with chosen implicit distributions automatically \cite{ml4co}. 

\textbf{Broader Impact} Our work shows {encouraging} economic and academic potential for incorporating machine learning to modern solvers. 
\textit{First}, when new presolvers are developed, the prior knowledge of them is usually limited. RL4Presolve helps us to learn the correlations of these new presolvers with existing ones and place them into right locations. 
For example, we employed several new presolvers in the commercial solver OptVerse, which are not in existing open-source LP solvers, for specific real-world applications. Following the steps in Section \ref{sec: exp-application}, we integrate them to OptVerse with better efficiency. 
\textit{Second}, in many real-world applications, we need to solve large-scale LPs repeatedly with limited time and hardware resources. Then, the improvement in solving LPs in these tasks usually brings enormous economic value. 
For example, decisions on supply chains usually require to be made within seconds and one percent optimization on the objective might bring millions of money saving. 
Therefore, the acceleration on industry-level benchmarks (e.g., benchmarks 1-3) can save time for other crucial downstream tasks and, eventually, save money in practice. 
\textit{Finally}, we note that routine optimization is a widespread challenge in software development \cite{ce, reformulate}. For example, modules like primal heuristic in MILP solvers \cite{scip} and pricing in LP solvers \cite{ct} may be optimized in similar manners. We hope our study can motivate more insightful research on these topics. 

\textbf{Limitations} There are still several limitations of this research that are remained for the future work. First, in this paper we mainly focus on the presolve process of linear programming. The optimization on the presolve process of other mathematical programming algorithms, e.g.,mixed integer linear programming, remains to be a future work. 
Second, similar to many previous researches \cite{ branch, heuristic, node-compare, cut}, the generalization ability to problems from totally different distributions, e.g., MIPLIB 2017, remains a challenge to be studied in the future. Finally, though we can extract rules from the learned polices to further optimize the hard-coded presolve routines in LP solvers, how to explain the learned rules to help us further understand the presolve process remains a future work.

\begin{table*}[t]
\caption{All presolvers integrated in Clp \cite{clp} sorted by their default execution orders. }
\centering
\begin{adjustbox}{width=0.8\textwidth}
\small
\begin{tabular}{lll}
\toprule\toprule
\multicolumn{1}{c}{Index} & Name             & Short description                                                                                                                                                                                                                              \\ \midrule
0                         & make fixed       & Fix variables if $\underline{x}_j=\overline{x}_j$.                                                                                                                                                                                \\ \midrule
1                         & test redundant   & \begin{tabular}[c]{@{}l@{}} Remove redundant constraints or tighten variable bounds\\  by comparing $L_i   = {\boldsymbol{a}}_{i,\cdot}^\top \underline{\mathbf{x}},\, U_i =  \boldsymbol{a}_{i,\cdot}^\top   \overline{\boldsymbol{x}}$ with $\underline{b}_i,\, \overline{b}_i$. \end{tabular}\\ \midrule
2                         & dupcol           & Fix redundant variables if $\boldsymbol{a}_{\cdot,j_1} = \alpha   \boldsymbol{a}_{\cdot,j_2}$.                                                                                                                                                  \\  \midrule
3                         & twoxtwo &  \begin{tabular}[c]{@{}l@{}} Remove constraints if some variables have two entries and\\ there are two entries with same other variable \cite{twoxtwo}.  \end{tabular}                                                                                                                                                                                                                                           \\ \midrule
4                         & duprow           & Remove redundant constraints if satisfy $\boldsymbol{a}_{i_1,\cdot} =   \alpha \boldsymbol{a}_{i_2,\cdot}$.                                                                                                                                     \\ \midrule
5                         & gubrow           & \begin{tabular}[c]{@{}l@{}} Reduce NNZ if the non-zero entries of $(\boldsymbol{a}_{i_1,\cdot} - \alpha   \boldsymbol{a}_{i_2,\cdot}) $ and \\ $ \boldsymbol{a}_{i_2,\cdot}$ are disjoint. \end{tabular}                                                                                                         \\ \midrule
6                         & implied free     & \begin{tabular}[c]{@{}l@{}} Remove redundant equations if $a_{i,k} \underline{x}_k + \sum_{j\neq   k}a_{i,j}\overline{x}_j\leq b_i$ \\and $a_{i,k} \overline{x}_k + \sum_{j\neq   k}a_{i,j}\underline{x}_j\geq b_i$.         \end{tabular}                                                 \\ \midrule
7                         & slack doubleton  &  Remove constraints like $\underline{b}_i \le a_{ij} x_j   \le \overline{b}_i$.                                                                                                                                                                                                                                              \\ \midrule
8                         & tighten action   & Fix variables if $c_j =0$ and ($\boldsymbol{a}_{\cdot, j} \geq \mathbf{0}   $ or $\boldsymbol{a}_{\cdot, j} \leq \mathbf{0} $).                                                                                                                 \\ \midrule
9                         & remove dual      & Tighten constraint and variable bounds by KKT conditions \cite{cvx}.                                                                                                                                                                                                                                               \\ \midrule
10                        & doubleton        & Fix variables for equations like $a_{i, j_1} x_{j_1} +   a_{i, j_2} x_{j_2} = b_i$.                                                                                                                                             \\ \midrule
11                        & tripleton        & Fix variables for equations like $a_{i, j_1} x_{j_1} +   a_{i, j_2} x_{j_2} + a_{i, j_3} x_{j_3} = b_i$.                                                                                                                        \\ \midrule
12                        & forcing          & Fix variables if $ \boldsymbol{a}_{i,\cdot}^\top   \underline{\mathbf{x}}\leq \underline{b}_i $ or $  \boldsymbol{a}_{i,\cdot}^\top   \overline{\mathbf{x}}\geq \overline{b}_i$.                                                                \\ \midrule
13                        & slack singleton  & \begin{tabular}[c]{@{}l@{}} Tighten the constraint bounds if some columns of the constraint\\ matrix $\boldsymbol{a}_{\cdot, j}$ have only one non-zero entry.  \end{tabular}                                                                                                                                         \\ \midrule
14                        & duprow3          & Remove redundant constraints $\boldsymbol{a}_{i_1,\cdot},   \boldsymbol{a}_{i_2,\cdot}, \boldsymbol{a}_{i_3,\cdot}$ are dependent.  \\  \bottomrule
\end{tabular}
\end{adjustbox}
\label{tab: presolvers}
\end{table*}

\begin{table*}[t]
\caption{The state feature vector used in RL4Presolve and their descriptions. }
\centering
\begin{adjustbox}{width=1.0\textwidth}
\small
\begin{tabular}{llll}
\toprule\toprule
Index & Feature type & Feature description & Normalization method\\  \midrule
0-3   &   equation degree &  Number of equations containing 1/2/3/4 non-zero entries.                   &    number of equations         \\
4     &    implied free   &  Number of equations that are implied free.        &       number of equations        \\
5-8   &  inequality degree  &  Number of inequalities containing 1/2/3/4 non-zero entries.    &   number of inequalities     \\
9     &  tighten    &  Number of variables that can be tightening. &   number of variables   \\
10-12 & statistics  &  Number of equations/inequalities/variables.   &   number of all constraints   \\
13-16 & forcing/redundant & Number of inequalities that are redundant or can be forced. &  number of all constraints  \\
17    &  NNZ  &  Current NNZ. &   constraints $\times$ variables \\  
18-35 & historical effect & The cumulative difference of features 0-17. &  the same as 0-17  \\  
36-50 & historical actions & The cumulative number of each executed presolver. &  not normalize \\
\bottomrule
\end{tabular}
\end{adjustbox}
\label{tab: features}
\end{table*}

\section{Further Discussions on the Approach} \label{sec: discussion-mdp}

\begin{figure*}[t]
\centering
\includegraphics[width=0.8\textwidth]{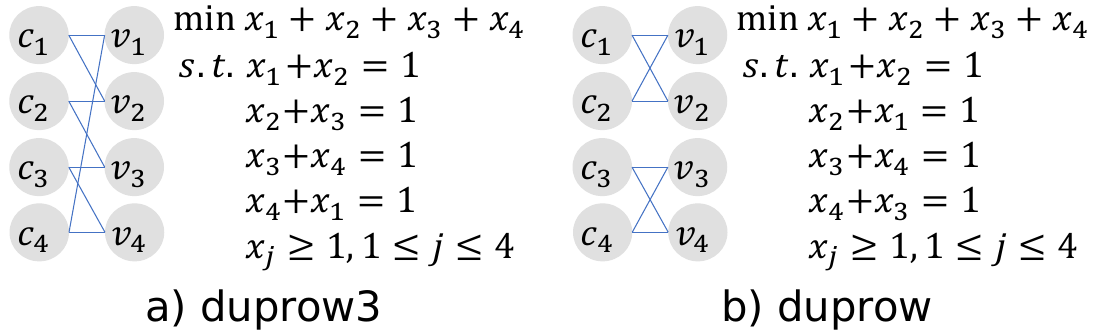}
\caption{An counterexample motivated from \citet{gnn-lp}. Suppose an LP solver (e.g., Clp) integrating both duprow and duprow3 in  presolve. Then, a GNN based presolve agent may have challenge to distinguish the two instances above due to its representation power.}
\label{fig: counterexample}
\end{figure*}

\textbf{Do Feature Vectors Contain Full Information for Presolve?} Though compressed feature vectors usually make the training process easier, they may loss information for the downstream task compared to original states \cite{rl-image}. 
In this paper, the feature vector defined in Table \ref{tab: features} does not contain the features for identifying dependent rows and constraints due to their high computational overhead\footnote{See Chapter 3 in \citet{ce} for detailed descriptions about how to calculate them.}, but these features are required for the presolvers 2-5,14 in Table \ref{tab: presolvers} \cite{ce}. 
To alleviate this limitation, we wonder whether GNNs can help to approximately extract these features from original bipartite graphs faster. 
However, motivated by the analysis in \citet{gnn-lp} of the representation power of GNNs on LP problems, we can find counterexamples to show that capturing the activation conditions for these dependencies is also challenging for GNNs (see Figure \ref{fig: counterexample} for a counterexample). 
In practice, we enhance the partially observable $\boldsymbol{s}$ by simply employing several statistical features about the whole presolve process (features 18-50 in Table \ref{tab: features}) to provide more historical information. 

\textbf{Does Time the Only Choice for Reward?} Note that the behavior of RL trained policies in the same environment can be totally different when we use different reward functions \cite{rlbook}. Thus, we would like to ask whether time is the only choice for reward? We use $r=-t$ in our MDP formulation as it is simple and directly related to (P1)-(P3), while another natural idea is to use $r=-\Delta \text{NNZ}$, as the NNZ reduction is one of the most important indicators about the presolve effect. However, this reward function has two severe limitations for tackling (P3). \textit{First}, $r=-\Delta \text{NNZ}$ is always non-negative. Thus,  RL agents tend to never stop presolve in such formulation. \textit{Second}, the total solving time is not always positively related to NNZ reduction (see the results on Set Covering in Table \ref{tab: routine-rule} for an example). 
Besides, using time directly as the reward reduces the expert knowledge required for problem formulation, making the RL objective in this task and our ultimate goal (i.e., improve the efficiency of solving LPs) to be end-to-end. 
Due to the above reasons, we finally decide to use only $r=-t$ as rewards.

\textbf{Why Only the One-Step Dependency is Modeled?} There are three reasons that why we only model the one-step dependency into adaptive action sequence. \textit{First}, the existence of one-step dependency for different presolvers is evident as demonstrated in Figure \ref{fig: p2}, while the existence of multi-step dependency is more unclear. \textit{Second}, the action space (i.e., the number of conditional probabilities to learn) grows exponentially with the step of dependency we consider. Thus, even two-step dependency makes the training process more difficult. \textit{Third}, the long-term dependency has already been included in states implicitly, as the effect of previous presolvers will first change the states and then influence the selection of presolvers at the next time step.

\section{Additional Experiments and Analysis}\label{app: exp}

\begin{figure*}[t]
    \centering
    \subfigure[Pie chart of NNZ reduction for all presolvers.]{
        \label{fig: p1-all}
        \includegraphics[width=0.8\textwidth]{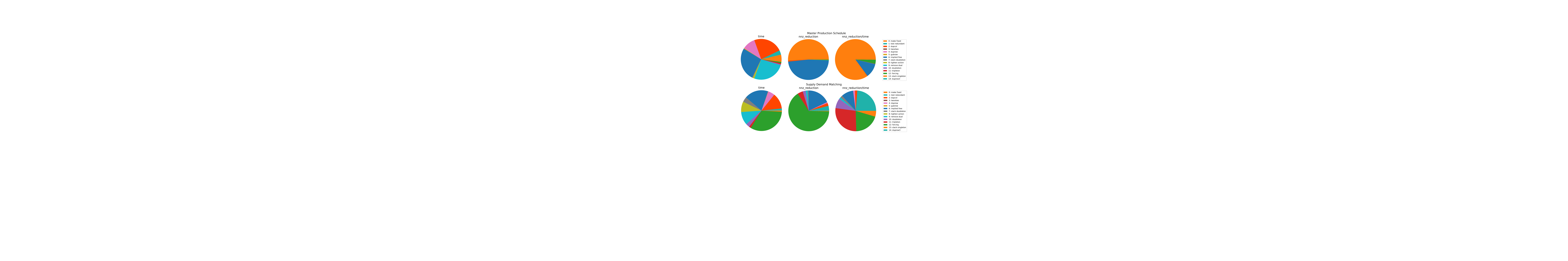}}
    \subfigure[Visualization for the presolve process.]{
        \label{fig: p3-all}
        \includegraphics[width=0.85\textwidth]{figures/process_all.pdf}}
\caption{Visualize (P1) and (P3) on the Master Production Schedule and Supply Demand Matching benchmarks. Results show that both (P1) and (P3) are critical in presolve routines.}
\label{fig: analysis-all}
\end{figure*}

\begin{figure*}[t]
\centering
\includegraphics[width=0.8\textwidth]{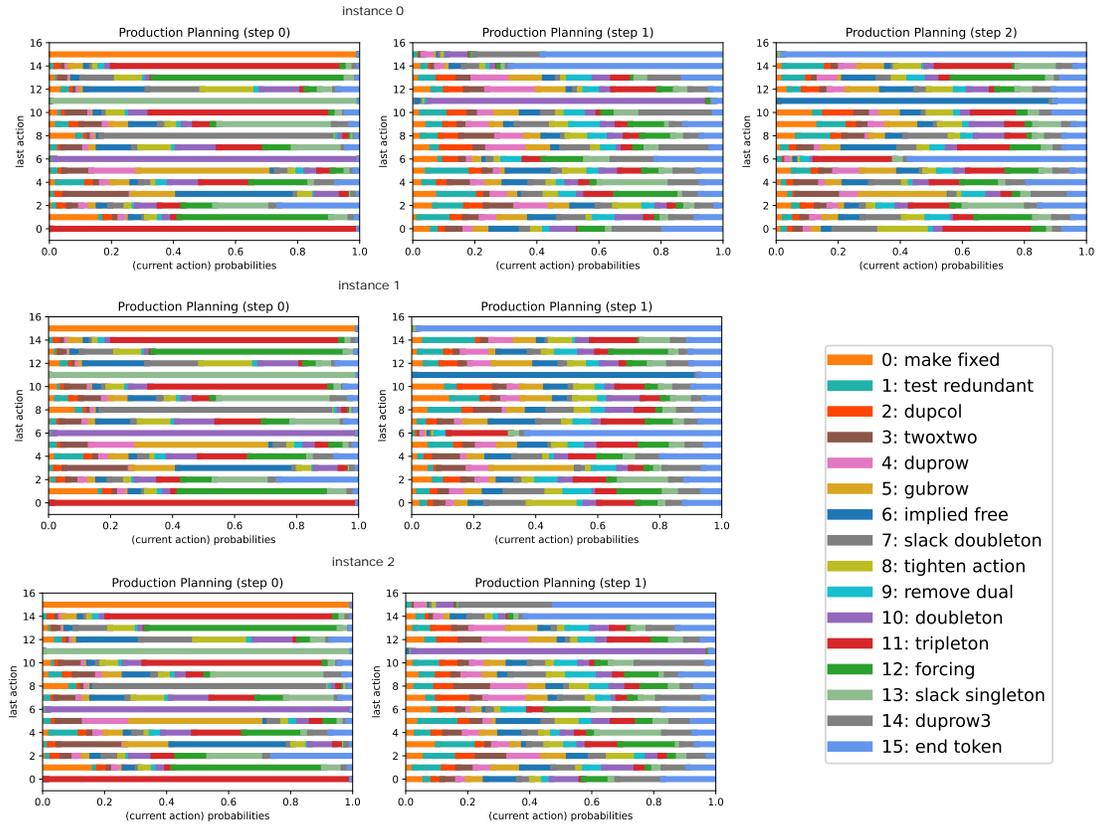}
\caption{Visualize the learned policies on three instances from the Production Planning benchmark. The results show that RL4Presolve tends to learn similar behaviors for instances from similar distribution, which motivates us to extract rules for each benchmark.}
\label{fig: policy-similarity}
\end{figure*}

\begin{figure*}[t]
    \centering
    \subfigure[Visualize the learned policies on Production Planning, Multicommodity Network Flow, and Set Covering.]{
        \label{fig: policy-visualize-1}
        \includegraphics[width=0.6\textwidth]{figures/action_probs_3.pdf}}
    \subfigure[Visualize the learned policy on MIRPLIB.]{
        \label{fig: policy-visualize-2}
        \includegraphics[width=1.0\textwidth]{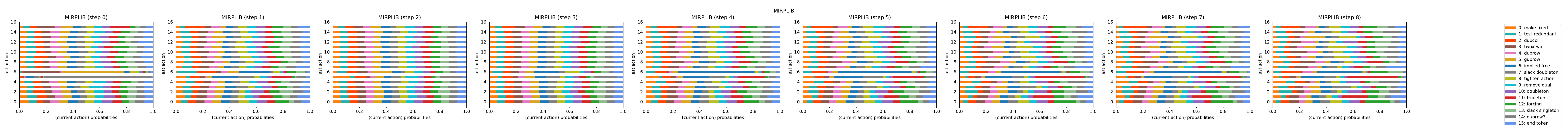}}
\caption{Visualize the learned policies on four different benchmarks to understand what RL4Presolve learned, which helps us extract rules and deploy to LP solvers. We observe that the one-step dependency learned by RL4Presolve matches the prior knowledge in Figure \ref{fig: p2}}.
\label{fig: policy-visualize}
\end{figure*}

\section{Implementation Details}\label{sec: more-implementations}

\begin{table*}[t]
\caption{All hyperparameters and network structures used in PPO in RL4Presolve.}
\centering
\begin{adjustbox}{width=0.8\textwidth}
\small
\begin{tabular}{lll}
\toprule\toprule
Parameter                                       & Value &   \\
\midrule
optimizer                                       & Adam  &  \\
learning rate $\beta$ (actor)                   & 1e-4  &   \\
learning rate  $\beta$ (critic)                 & 1e-4  &   \\
discount ($\gamma$)                             & 1.0   &   \\
training epoches $N$ per iteration              & 12    &    \\
number of collected samples $S$ per   iteration & 16    &   \\
minibatch size                                  & 16    &  \\
entropy coefficient                             & 1e-2  &   \\
parallel processes of sampling agents              & 4     &  \\
number of hidden layers (actor) & 2 & \\
number of hidden units per layer (actor) & 64 & \\
nonlinearity (actor) & Tanh & \\
number of hidden layers (critic) & 2 & \\
number of hidden units per layer (critic) & 64 & \\
nonlinearity (critic) & Tanh  & \\
\bottomrule
\end{tabular}
\end{adjustbox}
\label{tab: ppo-hyperparameters}
\end{table*}

\begin{table*}[t]
\caption{The size and the parameter for instance generation of all benchmarks. We report the  mean value with the standard deviation of all instances.}
\centering
\begin{adjustbox}{width=1.0\textwidth}
\small
\begin{tabular}{lllll}
\toprule\toprule
Benchmark                                       & Rows                                       & Columns               & NNZ                   & Parameters for instance generation                               \\ 
\midrule
{Master Production Schedule} & {9.58$e$5$\pm$(46.9\%)} & 2.29$e$6$\pm$(45.2\%) & 3.85$e$6$\pm$(45.0\%) & NA                                                               \\  \midrule
Production Planning                             & 3.16$e$5$\pm$(55.0\%)                      & 6.77$e$5$\pm$(56.1\%) & 1.68$e$6$\pm$(56.0\%) & NA                                                               \\ \midrule
Supply Demand Matching                          & 2.60$e$5$\pm$(26.0\%)                      & 5.18$e$5$\pm$(29.9\%) & 1.42$e$6$\pm$(18.7\%) & NA                                                               \\  \midrule
MIRPLIB                                         & 8.36$e$3$\pm$(53.2\%)                      & 2.55$e$4$\pm$(65.8\%) & 7.15$e$4$\pm$(67.5\%) & NA                                                               \\ \midrule
Facility Location                               & 2.00$e$3$\pm$(0.0\%)                       & 1.00$e$6$\pm$(0.0\%)  &  2.00$e$6$\pm$(0.0\%)  & \begin{tabular}[l]{@{}l@{}}number\_of\_customers=1000,\\ number\_of\_facilities=1000, ratio=2  \end{tabular}    \\ \midrule
Set Covering                                    & 3.00$e$4$\pm$(0.0\%)                       & 6.00$e$4$\pm$(0.0\%)  & 1.80$e$5$\pm$(0.0\%)  & \begin{tabular}[l]{@{}l@{}}nrow=3e4, ncol=6e4,\\ dens=1e-4, max\_coef=100  \end{tabular}                      \\ \midrule
Multicommodity Network Flow                     & 1.80$e$4$\pm$(0.3\%)                       & 8.94$e$5$\pm$(1.6\%)  & 1.34$e$6$\pm$(1.6\%)  & min\_n=max\_n=100                                                \\ \midrule
Generalized Network Flow                        & 5.01$e$4$\pm$(10.8\%)                      & 6.24$e$4$\pm$(10.3\%) & 1.25$e$5$\pm$(10.3\%) &   \begin{tabular}[l]{@{}l@{}} nodes=50000, nsorc=2500,\\ nsink=5000, dens=60000 \\all parameters are further\\randomized in the range of $\pm10\%$   \end{tabular}                                                  \\ \midrule
Production Planning (Large)                     & 1.36$e$6$\pm$(16.1\%)                      & 2.85$e$6$\pm$(17.1\%) & 7.10$e$6$\pm$(17.1\%) & NA                                                               \\ \midrule
Supply Demand Matching (Large)                  & 5.32$e$5$\pm$(36.2\%)                      & 9.24$e$5$\pm$(34.1\%) & 3.06$e$6$\pm$(36.5\%) & NA                                                               \\ \midrule
Multicommodity Network Flow (Large)             & 3.04$e$4$\pm$(0.2\%)                       & 1.96$e$6$\pm$(1.1\%)  & 2.93$e$6$\pm$(1.1\%)  & min\_n=max\_n=150                                                \\ \midrule
Facility Location (Large)                       & 3.00$e$3$\pm$(0.0\%)                       & 2.25$e$6$\pm$(0.0\%)  & 4.50$e$6$\pm$(0.0\%)  & \begin{tabular}[l]{@{}l@{}}number\_of\_customers=1500, \\ number\_of\_facilities=1500, ratio=2  \end{tabular}                                   \\
\bottomrule
\end{tabular}
\end{adjustbox}
\label{tab: benchmark-hyperparameters}
\end{table*}

\textbf{Hyperparameters in PPO} We report all hyperparameters used in PPO in RL4Presolve in Table \ref{tab: ppo-hyperparameters}. We normalize both the states and rewards with running mean and standard. We use independent actor and critic networks, i.e., no layers in these networks are shared with each other. We decrease the learning rates to half every $1000$ iterations for both the actor and the critic. 

\textbf{Details for Benchmarks}  
We report the range of rows, columns, and non-zero elements for all benchmarks and the corresponding parameters for the instance generation scripts in Table \ref{tab: benchmark-hyperparameters}.

\begin{table*}[t]
\caption{Total environment steps \cite{rlbook} and total training time (two NVIDIA Tesla V100 GPUs, four parallel processes of sampling agents in PPO) of each benchmark.}
\centering
\begin{adjustbox}{width=1.0\textwidth}
\small
\begin{tabular}{lcccc}
\toprule\toprule
                             & \multicolumn{2}{c}{Clp}                             & \multicolumn{2}{c}{OptVerse}                        \\ \midrule
Benchmark                    & Total environment steps & Total training time($h$) & Total environment steps & Total training time($h$) \\ \midrule
Master   Production Schedule & 1.0e4                   & 2.1                       & 1.0e4                   & 2.6                       \\
Production Planning          & 8.0e4                   & 16.7                      & 1.0e4                   & 3.3                       \\
Supply Demand Matching       & 3.0e4                   & 37.6                      & 1.5e4                   & 20.3                      \\
MIRPLIB                      & 1.0e4                   & 78.5                      & 1.0e4                   & 71.5                      \\
Facility   Location          & 0.5e4                   & 8.2                       & 0.5e4                   & 6.4                       \\
Set Covering                 & 0.5e4                   & 2.9                       & 0.5e4                   & 0.6                       \\
Multicommodity Network Flow  & 0.5e4                   & 2.4                       & 0.5e4                   & 1.8                       \\
Generalized Network Flow     & 30e4                    & 24.1                      & 10e4                    & 11.8                     \\ \bottomrule
\end{tabular}
\end{adjustbox}
\label{tab: env-step}
\end{table*}

\end{document}